\documentclass[10pt, conference, letterpaper]{IEEEtran}
\usepackage{amsmath,amsfonts}
\usepackage{array}
\usepackage[caption=false,font=normalsize,labelfont=sf,textfont=sf]{subfig}
\usepackage{textcomp}
\usepackage{stfloats}
\usepackage{url}
\usepackage{verbatim}
\usepackage{graphicx}
\usepackage{cite}
\usepackage{booktabs}
\usepackage{multirow}
\usepackage{graphicx}
\usepackage{bbm}
\usepackage{color}
\usepackage[T1]{fontenc}
\usepackage{aecompl}
\usepackage{dsfont}
\usepackage[linesnumbered,ruled,vlined]{algorithm2e}
\begin{document}
	
	\title{SpreadFGL: Edge-Client Collaborative Federated Graph Learning with Adaptive Neighbor Generation}
	
	\author{
		
		\IEEEauthorblockN{\small
			Luying Zhong$^{1,2}$, Yueyang Pi$^{1,2,3}$, Zheyi Chen$^{1,2,3,*}$,  Zhengxin Yu$^{4}$, Wang Miao$^{5}$, Xing Chen$^{1,2,3}$, and Geyong Min$^{6}$}
		\IEEEauthorblockA{\small
			$^{1}$College of Computer and Data Science, Fuzhou University, China\\
		}
		
		\IEEEauthorblockA{\small
			$^{2}$Fujian Provincial Key Laboratory of Network Computing and Intelligent Information Processing, Fuzhou University, China\\
		}
		
		\IEEEauthorblockA{\small
			$^{3}$Engineering Research Center of Big Data Intelligence, Ministry of Education, China\\
		}
		
		\IEEEauthorblockA{\small
			$^{4}$School of Computing and Communications, University of Lancaster, UK\\
		}
		
		\IEEEauthorblockA{\small
			$^{5}$School of Engineering, Computing and Mathematics, University of Plymouth, UK\\
		}
		
		\IEEEauthorblockA{\small
			$^{6}$Department of Computer Science, University of Exeter, UK\\
		}
		
		\IEEEauthorblockA{\small
			\{luyingzhongfzu@163.com, piyueyangcc@163.com, z.chen@fzu.edu.cn, z.yu8@lancaster.ac.uk, \\wang.miao@plymouth.ac.uk, chenxing@fzu.edu.cn, g.min@exeter.au.uk\}
		}
		
		\IEEEauthorblockA{\small
			$^*$Corresponding Author\\
		}
		
	}
	\maketitle
	\begin{abstract}
		Federated Graph Learning (FGL) has garnered widespread attention by enabling collaborative training on multiple clients for semi-supervised classification tasks.
		However, most existing FGL studies do not well consider the missing inter-client topology information in real-world scenarios, causing insufficient feature aggregation of multi-hop neighbor clients during model training.
		Moreover, the classic FGL commonly adopts the FedAvg but neglects the high training costs when the number of clients expands, resulting in the overload of a single edge server.
		To address these important challenges, we propose a novel FGL framework, named SpreadFGL, to promote the information flow in edge-client collaboration and extract more generalized potential relationships between clients.
		In SpreadFGL, an adaptive graph imputation generator incorporated with a versatile assessor is first designed to exploit the potential links between subgraphs, without sharing raw data.
		Next, a new negative sampling mechanism is developed to make SpreadFGL concentrate on more refined information in downstream tasks.
		To facilitate load balancing at the edge layer, SpreadFGL follows a distributed training manner that enables fast model convergence.
		Using real-world testbed and benchmark graph datasets, extensive experiments demonstrate the effectiveness of the proposed SpreadFGL.
		The results show that SpreadFGL achieves higher accuracy and faster convergence against state-of-the-art algorithms.
	\end{abstract}
	
	\begin{IEEEkeywords}
		Edge intelligence, federated graph learning, semi-supervised learning, neighbor generation.
	\end{IEEEkeywords}

	\section{Introduction}
	With powerful expressive capabilities, graphs \cite{ling2022malgraph} have been widely used to depict real-world application scenarios such as social network \cite{xie2021federated},  knowledge graph \cite{peng2021differentially}, and paper citation \cite{chen2022graph}. 
	In the area of graph learning, the emerging Graph Neural Networks (GNNs) have gained significant attention due to their exceptional performance in dealing with graph-related tasks. 
	GNNs efficiently utilize the feature propagation by employing multiple graph convolutional layers for node classification tasks, where the structural knowledge is distilled into discriminative representations from complex graph-orient data in diverse domains such as prediction modeling \cite{geyer2019deeptma}, malware detection \cite{ling2022malgraph}, and resource allocation \cite{han2021tailored}. %
	Commonly, the training performance of GNNs depends on the substantial graph data distributed among clients. 
	However, due to privacy and overhead concerns, it is impractical to assemble graph data from all clients for GNN training.
	
	\begin{figure}[t]
		\centering
		\begin{center}
			\includegraphics*[width=1\linewidth]{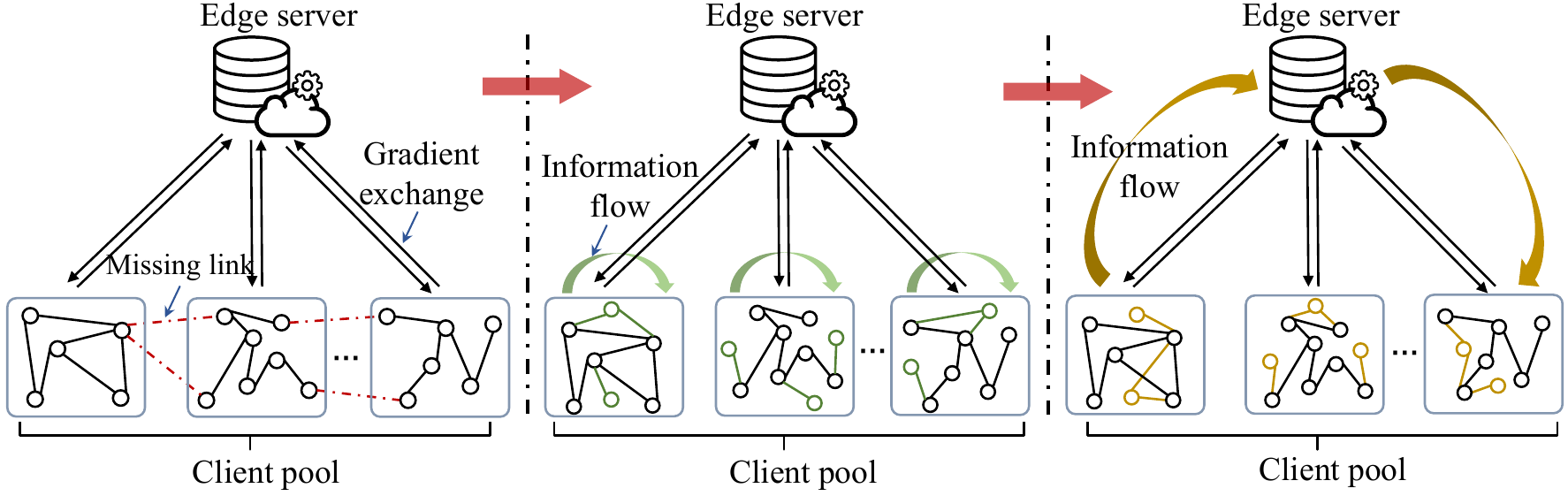}
			\vspace{-0.5cm}
			\caption{Comparison between the classic FGL and the FedGL designed in the proposed SpreadFGL.
				In Fig. \ref{fig:figure_1} (left), the FGL \cite{scardapane2020distributed} does not consider the inter-links between clients, causing insufficient feature propagation of multi-hop neighbors.
				In Fig. \ref{fig:figure_1} (middle), the FGL \cite{zhang2021subgraph} solely infers the missing links by local subgraphs but ignores the meaningful information in neighbor clients.
				In Fig. \ref{fig:figure_1} (right), the proposed FedGL utilizes the globally-shared information among clients, thereby extracting important cross-subgraph links for classification tasks.
			}
			\label{fig:figure_1}
			\vspace{-0.7cm}
		\end{center}
	\end{figure}
	
	Following a distributed training mode, Federated Graph Learning (FGL) aims to deal with the problem of graph data island by promoting cooperative training among multiple clients \cite{li2021sample}.
	To protect privacy, the FGL offers generalized graph mining models over distributed subgraphs without sharing raw data \cite{scardapane2020distributed}.
	Many studies have verified the feasibility of FGL in various domains such as transportation \cite{zhang2021fastgnn}, computer vision \cite{chen2021fedgraph}, and edge intelligence \cite{wang2023graph}. 
	Recently, some studies also adopted FGL-based frameworks for semi-supervised classification tasks \cite{zhang2021subgraph}, \cite{wang2022graphfl}.
	These approaches typically join an edge server with multiple clients to train a globally-shared classifier for downstream tasks, where the clients and edge server undertake local updating and global aggregation, respectively.
	
	In real-world FGL application scenarios, there are potential links between the subgraphs of a client and others since these subgraphs contain significant information about neighbor clients \cite{scardapane2020distributed}.
	However, previous FGL-related studies \cite{DBLP:conf/aaai/0001CBAA22}, \cite{yuan2022fedstn} overlooked such important links among clients, as shown in Fig. \ref{fig:figure_1} (left). 
	This oversight results in the insufficient feature propagation of multi-hop neighbors during local model training, ultimately leading to degraded performance in classification tasks. 
	To explore the potential links among clients, some prior studies inferred the missing links in subgraphs, as shown in Fig. \ref{fig:figure_1} (middle).
	For example, Zhang \textit{et al.} \cite{zhang2021subgraph} employed linear predictors on local models to conduct the missing links in subgraphs.
	However, the potential links rely solely on local clients, disregarding meaningful information from neighbor clients.
	Therefore, the features implied in the generated links may be incomplete and infeasible to recover the cross-client information.
	Moreover, most existing studies \cite{zhang2021subgraph}, \cite{caldarola2021cluster} commonly adopted the classic FedAvg algorithm \cite{DBLP:conf/aistats/McMahanMRHA17}, neglecting the high training costs when the number of clients continues to expand, which leads to a serious single-point overload problem.
	
	To address these essential challenges, we propose FedGL, an improved centralized FGL framework, to explore potential cross-subgraph links by leveraging the global information flow.
	As illustrated in Fig. \ref{fig:figure_1} (right), we consider the edge server as an intermediary to facilitate the flow of global information, thereby enhancing communication among different clients and fostering the collaborative knowledge integration of their subgraphs.
	Thus, the proposed FedGL is able to extract unbiased and generalized missing links through collaboration among the edge server and clients.
	Furthermore, we extend the FedGL to a multi-edge collaborative scenario and propose the SpreadFGL to efficiently handle the load balancing issue at the edge layer. 
	The main contributions of this paper are summarized as follows.
	\begin{itemize}
		\item We propose an improved centralized FGL framework, named FedGL.
		In FedGL, GNNs are utilized as the node classifiers in clients for semi-supervised classification tasks, ensuring effective feature propagation.
		\item We design an adaptive graph imputation generator to explore generalized potential cross-subgraph links,  referring to the globally-shared topology graph at the edge layer.
		\item We develop a new versatile assessor that incorporates a negative sampling mechanism to supervise the process of generating subgraphs, where the discriminate features are constructed by autoencoder.
		Thus, we can focus on more refined features that are beneficial to classification tasks.
		\item We propose a novel distributed FGL framework, named SpreadFGL, which extends the FedGL to a multi-edge scenario.
		In SpreadFGL, the neighbor edge servers collaboratively conduct model training with a well-designed distributed loss function, enabling efficient extraction of the potential links between subgraphs.
		Thus, the SpreadFGL facilitates faster model convergence and better load balancing at the edge layer.
		\item Using real-world testbed and benchmark graph datasets, extensive experiments are conducted to demonstrate the superiority of the proposed SpreadFGL.
		The results show that the SpreadFGL outperforms state-of-the-art algorithms from the perspectives of model accuracy and convergence speed.
	\end{itemize}

	The rest of this paper is organized as follows.
	Section \ref{related_work} reviews the related work of GNNs and FGL.
	Section \ref{method} elaborates the proposed FedGL and SpreadFGL.
	Section \ref{experiment} evaluates the proposed frameworks via extensive experiments.
	Section \ref{conclusion} concludes this paper.
	
	\section{Related Work}\label{related_work}
	\subsection{Graph Neural Networks}
	Graph Neural Networks \cite{wu2020comprehensive} have drawn considerable attention in recent years due to their remarkable capabilities.
	As an emerging technique in semi-supervised learning, GNNs can achieve accurate node classification for massive unlabeled nodes by training scarce labeled data.
	Considering the advanced ability in modeling graph structures, GNNs have derived several variants such as Graph Convolutional Networks (GCNs) \cite{DBLP:conf/iclr/KipfW17}, Graph Attention Networks (GAT) \cite{DBLP:conf/iclr/VelickovicCCRLB18}, and GraphSAGE \cite{DBLP:conf/nips/HamiltonYL17}.
	For example, GCNs conduct the operations of neural networks on graph topology, which have been widely used in semi-supervised learning tasks.
	The inference vector of the node $u$ on the $(l+1)$-th GCN layer is defined as
	\vspace{-0.1cm}
	\begin{equation}
		\begin{split}
			\mathbf{h}_u^{(l+1)}=\sigma\left\{\mathop{\mbox{AGG}}\left(\mathbf{h}_v^{(l)},e_{uv}\right)| \forall v \in \mathcal{V}\right\},\\
		\end{split}
		\label{gcn}
	\end{equation}
	where $\mathbf{h}_v^{(l)}$ is the vector of the node $u$ in the $l$-th GCN layer.
	$e_{uv}$ indicates the link between the node $u$ and $v$.
	$\mbox{AGG}(\cdot)$ is an aggregator function used to integrate the neighbor features of node $u$ via $e_{uv}$.
	And $\sigma(\cdot)$ is a non-linear activation function.
	
	The GAT incorporates GCNs with attention mechanisms to adaptively assign the weights $\alpha_{uv}^{(l+1)}$ for the neighbors of the node $u$, and the inference vector is defined as
	\vspace{-0.1cm}
	\begin{equation}
		\begin{split}
			\mathbf{h}_u^{(l+1)}=\sigma\left\{\mathop{\mbox{AGG}}\left(\alpha_{uv}^{(l+1)}\mathbf{h}_v^{(l)},e_{uv}\right)| \forall v \in \mathcal{V}\right\}.\\
		\end{split}
		\label{gat}
	\end{equation}
	
	The GraphSAGE aggregates node features by sampling from neighbor nodes and the inference vector is defined as
	\vspace{-0.1cm}
	\begin{equation}
		\begin{split}
			\mathbf{h}_u^{(l+1)}=\sigma\left\{\mathbf{h}_u^{(l)} \Vert \mathop{\mbox{AGG}}\left(\mathbf{h}_v^{(l)},e_{uv}\right)| \forall v \in \mathcal{V}\right\},\\
		\end{split}
		\label{graphsage}
	\end{equation}
	where $\Vert$ denotes the concatenation operation.
	
	Many scholar have contributed to GNN-based semi-supervised learning.
	For instance, Wang \textit{et al.} \cite{wang2020gcn} proposed a GCN framework that conducted feature propagation in topology and node spaces, aiming to promote the fusion of graphs and features.
	Zhong \textit{et al.} \cite{zhong2023contrastive} designed a contrastive GCN framework with a generative adjacency matrix to explore the topology correlations for downstream tasks.
	Sun \textit{et al.} \cite{sun2020multi} adopted a multi-stage GCN-based framework that employed self-supervised learning to compensate for the limit labeled signal.
	Although the existing studies have gained great success in centralized semi-supervised learning, they did not well consider the discrete distribution of graph data that occurred in real-world scenarios.
	It should be noted that the message passing between subgraphs will be blocked if the cross-subgraph connections are missing.
	This seriously violates the propagation of GNNs in multi-hop neighbors and leads to unsatisfactory performance.
	Therefore, there is an urgent need to study the restoration of missing cross-subgraph links for better handling the semi-supervised node classification.
	
	\begin{figure*}[t]
		\centering
		\begin{center}
			\includegraphics*[width=1\linewidth]{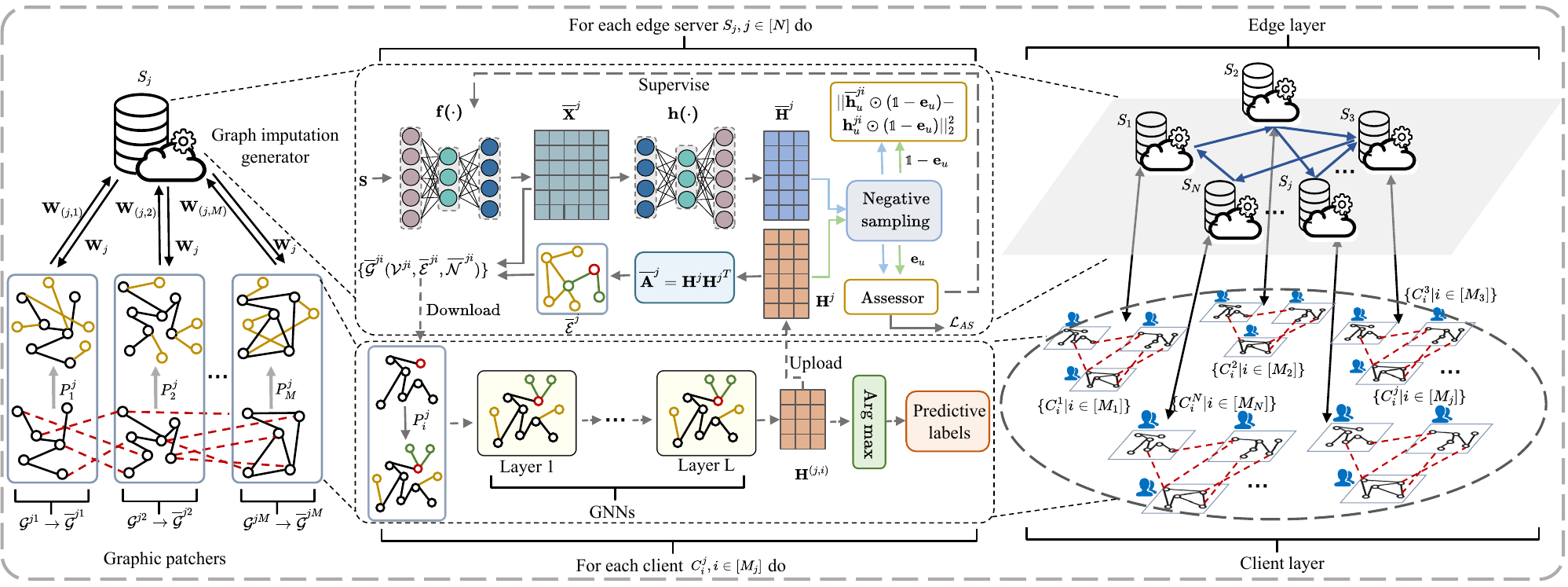}
			\vspace{-0.5cm}
			\caption{Overview of the proposed SpreadFGL.
				The SpreadFGL targets a distributed scenario that consists of multiple edge servers and clients.	
				At the edge layer, the autoencoder is employed to explore potential global features of the covered clients, and then the versatile assessor is combined with a negative sampling mechanism to supervise refined information, where model parameters transmission is permitted between neighbor edge servers.
				At the client layer, GNNs are used as local node classifiers for downstream tasks, and then graphic patchers are employed to repair subgraphs and missing cross-subgraph links.
			}
			\label{fig:figure_3}
		\end{center}
		\vspace{-0.4cm}
	\end{figure*}
	\subsection{Federated Graph Learning}
	Federated Graph Learning (FGL) \cite{yuan2022fedstn}, \cite{tan2023federated}, \cite{wang2022federatedscope} has emerged as a captivating topic in recent years.
	Different from the classic GNN that relies on centralized feature propagation across the entire graph, FGL enables distributed clients to collectively maintain a globally-shared model through gradient aggregation. 
	Many efforts have contributed to this topic.
	For instance, He \textit{et al.} \cite{he2021fedgraphnn} proposed a graph-level scheme that distributed graph datasets across multiple clients, catering to various downstream tasks.
	Wu \textit{et al.} \cite{wu2021fedgnn} designed an FGL framework for recommendation tasks, where subgraphs contain overlapped items.
	Xie \textit{et al.} \cite{xie2021federated} developed an FGL-based framework to mitigate the heterogeneity among features and graphs. 
	They employed clustering techniques to aggregate clients based on the GNN gradients, aiming to enhance the collaboration efficiency of federated learning.
	
	However, the above studies overlooked the pervasive missing links between clients happened in real-world scenarios, which may cause undesired performance in downstream tasks.
	To the best of our knowledge, few studies well considered and tackled the problem of missing cross-subgraph links. 
	Zhang \textit{et al.} \cite{zhang2021subgraph} utilized a local linear predictor to explore the potential relationships between clients according to the local subgraph structure.
	However, the cross-subgraph relationships rely on important information from neighbor clients, which makes it hard to find the potential links only using local subgraphs, thereby leading to inefficient recovery of cross-client information.
	Moreover, prior studies commonly adopted the classic FedAvg for training, ignoring the overload of a single node (e.g., edge server) especially when the number of clients expands.
	
	\section{The Proposed SpreadFGL} \label{method}
	In this section, we consider the typical FGL scenario with distributed graph datasets.
	Based on this setting, we first propose an improved centralized FGL framework, named FedGL.
	Next, we extend the FedGL to the scenario of multi-edge collaboration and propose a novel distributed FGL framework, named SpreadFGL.
	Specifically, Fig. \ref{fig:figure_3} provides a detailed illustration of the proposed SpreadFGL.

	\subsection{Overview and Motivation}
	A graph dataset is denoted as $\mathcal{D}\left(\mathcal{G}, \mathbf{Y}\right)$, where $\mathcal{G}=(\mathcal{V},
	\mathcal{E}, \mathbf{X})$ is a global graph.
	$\mathcal{V}$ is the node set, where $|\mathcal{V}|=n$.
	$\mathcal{E}=\left\{e_{uv}\right\}$ is the edge set that stores the link relationship between the nodes $u$ and $v$, where $\forall u,v \in \mathcal{V}$. 
	$\mathbf{X} \in \mathbb{R}^{n\times d}$ indicates the node feature matrix, where $\mathbf{x}_i \in \mathbb{R}^d$ is the feature vector of the $i$-th node.
	$\mathbf{Y}=[0,1]\in \mathbb{R}^{n\times c}$ is the label matrix, where $c$ is the number of classes.
	Considering that there are $N$ edge servers and $M$ clients.
	The edge server $S_j$ covers $M_j$ local clients $\{C^{j}_i|i\in [M_j]\}$ to conduct the FGL training, where $\sum_{j=1}^{N}M_j=M$.
	The client $C^{j}_i$ owns the part samples of the graph dataset, denoted by $\mathcal{D}^j_{i}\left\{\mathcal{G}^{ji}, \mathbf{Y}^{ji}\right\}$, where $\mathcal{G}^{ji}=\left(\mathcal{V}^{ji}, \mathcal{E}^{ji}, \mathbf{X}^{ji}\right)$ is a local subgraph and $\mathbf{Y}^{ji}$ is the sub-label matrix of nodes $\mathcal{V}^{ji}$.
	To simulate the real-world scenario of missing links between clients, we consider that there are no shared nodes and connected links among clients, formulated by $\mathcal{V}^{ji} \cap \mathcal{V}^{\hat{j}r} = \emptyset$, where $\forall i,r \in [M_j]$ and $i \neq r$ if $j=\hat{j}$, and $\forall i\in [M_j], \forall r\in [M_{\hat{j}}]$ if $j\neq \hat{j}$.
	The subgraphs of all clients form the complete graph, defined by $\sum_{j=1}^N\sum_{i=1}^{M_j}{|\mathcal{V}^{ji}|}=n$.
	Thus, there is no link between any two clients, and a client cannot directly retrieve the node features from another client.
	
	Based on the above scenario, the client $C_i^j$ owns a local node classifier $F^j_{i}$ and graphic patcher $P^j_{i}$, and all clients can jointly learn graph representations for semi-supervised node classification.
	Generally, the proposed SpreadFGL aims to conduct collaborative learning on independent subgraphs across all clients, prioritizing the privacy of raw data.
	Therefore, the SpreadFGL obtains the global node classifiers $\left\{F_j|j\in [N]\right\}$ parameterized by $\{\mathbf{W}_{j}|j\in [N]\}$ in the edge servers for downstream tasks.
	With this consideration, we formulate the optimization problem as minimizing the aggregated risks to find the optimal weights $\{\mathbf{W}_{j}|j\in [N]\}$, defined as
	\vspace{-0.1cm}
	\begin{equation}
		\begin{split}
			\sum_{j=1}^N\min_{\mathbf{W}_{j}} \mathcal{R}_j\left(F_j(\mathbf{W}_{j})\right)=\sum_{j=1}^N\min_{\mathbf{W}_{(j,i)}} \frac{1}{M_j}\sum_{i=1}^{M_j}\mathcal{R}^{j}_i\left(F^{j}_i(\mathbf{W}_{(j,i)})\right), \\
		\end{split}
		\label{loss1}
		\vspace{-0.1cm}
	\end{equation}
	where $\mathbf{W}_{(j,i)}$ is the learnable weights of local node classifier $F^{j}_i$.
	$\mathcal{R}_j$ is the loss function of the global node classifier $F_j$.
	And $\mathcal{R}^{j}_i$ is the loss function of the $i$-th client that is used to measure the local empirical risk, 
	
	\begin{small}
		\begin{equation}
			\begin{split}
				\mathcal{R}^{j}_i\left(F^{j}_i(\mathbf{W}_{(j,i)})\right)=\hspace{-0.1cm}\frac{1}{|\mathcal{V}^{ji}_t|}\hspace{-0.15cm}\sum_{v \in \mathcal{V}^{ji}_t}\hspace{-0.1cm}\mathcal{R}^j_i\left(\mathbf{W}_{j};F^j_i(\mathcal{G}^{ji}(v)), y^{ji}_v\right),\\
			\end{split}
			\label{loss2}
			\vspace{-0.4cm}
		\end{equation}
	\end{small}
	\hspace{-0.15cm}where $\mathcal{V}^{ji}_t \subseteq \mathcal{V}^{ji}$ is the labeled training set in the $i$-th client, and $y_v^{ji}$ is the ground truth of node $v$ in the $i$-th client.
	
	\subsection{FedGL: Centralized Federated Graph Learning}
	Since clients cannot directly capture cross-subgraph links that contain important neighbor information, the feature propagation from higher-order neighbors becomes inadequate, resulting in degraded classification performance.
	Therefore, it is crucial to explore the potential topology links among clients.
	To achieve this goal, we propose an improved centralized FGL framework, named FedGL.
	In FedGL, we consider an edge server to communicate with $M$ clients.
	The FedGL leverages the edge server $S_j$ as an intermediary to facilitate the information flow among clients, where $S_j$ covers all clients, denoted by $M_j=M$.
	Specifically, we incorporate a graph imputation generator to construct learnable links, thereby generating the latent links between subgraphs.
	To enhance feature propagation in local tasks and facilitate subsequent inference with the global model, we employ a $L$-layer GNN model with the local node classifier $F^j_i$, defined as
	\vspace{-0.1cm}
	\begin{equation}
		\begin{split}
			\mathbf{H}^{(j,i)}=\mbox{GNNconv}_{\mathbf{W}_{(j,i)}}(\mathcal{E}^{ji}, \mathbf{X}^{ji}),\\
		\end{split}
		\label{Conv}
	\end{equation}
	where $\mbox{GNNconv}(\cdot)$ is a GNN model and $\mathbf{H}^{(j,i)}$ indicates the GNN output of the $i$-th client covered by $S_j$.
	The feature propagation of the $(l+1)$-th layer is given in Eq. \eqref{graphsage}.
	Moreover, the Cross-Entropy loss function is adopted for the $i$-th client covered by $S_j$ in the downstream tasks, defined as
	\vspace{-0.1cm}
	\begin{equation}
		\begin{split}
			\mathcal{L}_{F_i^j}=\mathcal{R}^j_i\left(F^j_i(\mathbf{W}_{(j,i)})\right)=-\sum_{u=1}^{|\mathcal{V}_t^{ji}|}\sum_{r=1}^c\mathbf{Y}_{ur}^{ji}ln\mathbf{H}^{(j,i)},\\
		\end{split}
		\label{loss3}
	\end{equation}
	where $\mathbf{Y}^{ji}_u$ is the inference vector of the node $u$ conducted by local training.
	
	For every edge-client communication in FedGL, each client parallelly trains the local node classifier $F^j_i$ parameterized by $\mathbf{W}_{(j,i)}$ in local training rounds, formulated as
	\vspace{-0.1cm}
	\begin{equation}
		\begin{split}
			\mathbf{W}_{(j,i)}^{t+1}=\mathbf{W}^{t}_{(j,i)}-\alpha\nabla\mathcal{R}_i\left(F_i(\mathbf{W}_{(j,i)}^t)\right),\\
		\end{split}
		\label{update1}
	\end{equation}
	where $\alpha$ is the learning rate.
	$t\in [T_l-1]$ indicates the local training rounds.
	
	After local training, $S_j$ aggregates local parameters $\left\{\mathbf{W}_{(j,i)}|i\in [M_j]\right\}$ to update global ones $\mathbf{W}_j$, and then broadcasts $\mathbf{W}_j$ to all clients at each edge-client communication.
	
	\subsection{Graph Imputation Generator with Versatile Assessor}
	To capture the potential cross-subgraph links, we design a graph imputation generator and incorporate it with a versatile assessor to explore a learnable potential graph $\overline{\mathcal{G}}^j=(\mathcal{V}^j, \overline{\mathcal{E}}^j, \overline{\mathbf{X}}^j)$ for mending the cross-subgraph links.
	
	\textbf{Graph Imputation Generator.}
	To construct the globally-shared information without revealing raw data, the clients upload the processed embeddings $\left\{\mathbf{H}^{(j,i)}|i\in [M_j]\right\}$  to the edge server at every $\mathcal{K}$ intervals of edge-client communication, where the original linked nodes remain proximate in the low-dimensional space.
	Next, the graph imputation generator performs the fusion on the processed embeddings to obtain the globally-shared information $\mathbf{H}^j \in \mathbb{R}^{|\mathcal{V}^j| \times c}$, where $\mathcal{V}^j$ is the number of all clients covered by $S_j$.
	Based on this, $\mathbf{H}^j$ is denoted as
	\vspace{-0.3cm}
	\begin{equation}
		\begin{split}
			\mathbf{H}^j=\left[\mathbf{H}^{(j,1)} \Vert \cdots \Vert \mathbf{H}^{(j,M_j)}\right]. \\
		\end{split}
		\label{cat}
	\end{equation}
	
	In real-world application scenarios of FGL, it is possible for each node in clients to own potential cross-subgraph links, and it may be insufficient for clients to propagate features in multi-hop neighbors if missing these cross-subgraph links.
	In response to this problem, the graph imputation generator utilizes the distance to evaluate the node similarity and construct the global topology graph, referred to $\overline{A}^j=\mathbf{H}^j{\mathbf{H}^j}^T$.
	Next, $k$ most similar nodes are selected from this topology graph as potential cross-subgraph links, denoted by the set $\overline{\mathcal{E}}^j$.
	To generate the potential feature vectors $\overline{\mathbf{X}}^j$ under the guidance of the globally-shared information, an autoencoder parameterized by $\Phi_{AE}$ is used to explore overcomplete underlying representations from $\mathbf{H}^j$.
	Furthermore, to guarantee data privacy, the random noisy vector $\mathbf{S}$ is input to the autoencoder, and thus the output of the autoencoder is reconstructed as $\overline{\mathbf{H}}^j = h\left(f(\mathbf{\mathbf{S}})\right)$, where $f(\cdot)$ and $h(\cdot)$ are the encoder and decoder, respectively.
	It is noted that $\overline{\mathbf{X}}^j=f(\mathbf{S})$ indicates the potential features expected to be extracted by the encoder.
	With the autoencoder, the random noisy vector is mapped to the same dimension as $\mathbf{H}^j$, and the output of the $(l+1)$-th layer is defined as
	\vspace{-0.1cm}
	\begin{equation}
		\begin{split}
			\overline{\mathbf{H}}^{(j,l+1)}=\sigma\left(\overline{\mathbf{H}}^{(j,l)}\mathbf{W}^{(j,l+1)}_a+\mathbf{b}^{(j,l+1)}_a\right), \\
		\end{split}
		\label{autoencoder}
		\vspace{-0.1cm}
	\end{equation}
	where $\mathbf{W}_a^{(j,l+1)} \in \mathbb{R}^{d_{l}\times d_{l+1}}$ and $\mathbf{b}_a^{(j,l+1)} \in \mathbb{R}^{d_l}$ are the layer-specific weights and biases, respectively.
	$\sigma(\cdot)$ denotes the activation function. 
	
	\textbf{Versatile Assessor.}
	Since the conditional distribution of $\overline{\mathbf{H}}^j$ relies on $\overline{\mathbf{X}}^j$ and is independent of $\mathbf{S}$, $\left\{\mathbf{S}\rightarrow \overline{\mathbf{X}}^j\rightarrow \overline{\mathbf{H}}^j\right\}$ in the autoencoder follows the Markov principle.
	Therefore, we design a versatile assessor parameterized by $\Phi_{AS}$ to supervise the quality of reconstruction data from the decoder, aiming to extract the expected underlying features $\overline{\mathbf{X}}^j$ tailored for node classification.
	Considering the diversity of datasets, the assessor should be trainable to fit in specific tasks.
	Thus, the assessor adopts a fully-connected neural network to evaluate $\overline{\mathbf{H}}^j$.
	Concretely, the assessor takes the reconstructed globally-shared information $\overline{\mathbf{H}}^j$ as input in the form of a value, which is positively correlated with the quality evaluation of the reconstructed data.
	Hence, the autoencoder tends to obtain a higher value under the supervision of the assessor and extract more valid global information.
	Specifically, the loss function of the autoencoder is defined as
	\vspace{-0.1cm}
	\begin{equation}
		\begin{split}
			\hat{\mathcal{L}}_{AE} &=-\sum\nolimits_u\mathbb{E}_{p(\overline{\mathbf{h}}^j_u|\forall u \in \mathcal{V}^j)}\left(Assor\left(\overline{\mathbf{h}}^j_u\right)\right), \\
			&=\frac{1}{|\mathcal{V}^j|}\sum\nolimits_u\log\left(1-Assor\left(\overline{\mathbf{h}}^j_u\right)\right), \\
		\end{split}
		\label{autoencoder_loss1}
	\end{equation}
	where $\mathbb{E}_p{(\cdot)}$ is the expectation of the variables in $p(\cdot)$, and $p(\overline{\mathbf{h}}^j_u|\forall u \in \mathcal{V}^j)$ indicates $\overline{\mathbf{h}}^j_u$ sampled from the distribution of $\overline{\mathbf{H}}^j$.
	$Assor(\cdot)$ is the assessor that evaluates the constructed global information.
	To distinguish the original and reconstructed global data, we regard the globally-shared information as the criterion and train the assessor to assign higher scores.
	In contrast, the assessor is trained to assign lower scores with the reconstructed global information.
	Therefore, the assessor is able to guide the autoencoder to evolve more discriminative representations of latent features.
	Specifically, the loss function of the assessor is defined as
	
	\vspace{-0.5cm}
	\begin{small}
		\begin{equation}
			\begin{split}
				\hat{\mathcal{L}}_{AS} &=-\sum\nolimits_u\big[\mathbb{E}_{p(\mathbf{h}^j_u|\forall u \in \mathcal{V}^j)}\left(Assor\left(\mathbf{h}^j_u \right)\right) \\ &+\mathbb{E}_{p(\overline{\mathbf{h}}^j_u|\forall u \in \mathcal{V}^j)}\left(1-Assor\left(\overline{\mathbf{h}}^j_u\right)\right) \big] \\
				&=\frac{1}{|\mathcal{V}^j|}\sum\nolimits_u\left[\log\left(1-Assor\left(\mathbf{h}^j_u\right)\right) + \log\left(Assor\left(\overline{\mathbf{h}}^j_u\right)\right)\right], \\
			\end{split}
			\label{assessor_loss1}
			\vspace{-0.3cm}
		\end{equation}
	\end{small}
	\hspace{-0.15cm}where $p(\mathbf{h}^j_u|\forall u \in \mathcal{V}^j)$ denotes $\mathbf{h}^j_u$ sampled from the distribution of $\mathbf{H}^j$.
	
	The training processes of the autoencoder and assessor are performed simultaneously, where the assessor guides the autoencoder to learn more discriminative reconstructed data and potential features through back-propagation.
	
	\subsection{Negative Sampling and Graph Fixing}
	\textbf{Negative Sampling.}
	To extract more refined potential features, we develop a negative sampling mechanism to concentrate on the pertinent information for node classification.
	Based on the proposed versatile assessor, we first set a threshold $\theta\in (0,1)$ in every training iteration of the autoencoder and select the attributes in $\mathbf{h}^j_u$ that are less than $\theta$.
	These attributes are deemed as negative and their feedbacks from the assessor are 0.
	Next, the zero-regularization is used to process these negatives, and thus both the autoencoder and the assessor can spotlight the representations that are meaningful for downstream tasks.
	Hence, the loss function of the assessor is updated and redefined as
	\vspace{-0.1cm}
	\begin{equation}
		\begin{split}
			\mathcal{L}_{AS}&=\frac{1}{|\mathcal{V}^j|}\sum\nolimits_u\bigl[\log\left(1-Assor\left(\mathbf{h}^j_u\odot \mathbf{e}_u\right)\right) \\
			&+ \log\left(Assor\left(\overline{\mathbf{h}}^j_u\odot \mathbf{e}_u\right)\right)\bigl], \\
		\end{split}
		\label{assessor_loss2}
	\end{equation}
	where $\mathbf{e}_u$ is a $c$-dimensional vector that judges whether $h^j_{ui}\in \mathbf{h}^j_u$ is higher than $\theta$ ($e_{ui}=1$) or not ($e_{ui}=0$).
	$\odot$ is the element-wise multiplication.
	
	Correspondingly, the loss function of the autoencoder is updated and redefined as
	\begin{equation}
		\begin{split}
			\mathcal{L}_{AE}&=\frac{1}{|\mathcal{V}^j|}\sum\nolimits_u\bigl[\log\left(1-Assor\left(\overline{\mathbf{h}}^j_u\odot \mathbf{e}_u\right)\right) \\
			&+\Vert\mathbf{h}^j_{u}\odot (\mathds{1}-\mathbf{e}_{u})-\overline{\mathbf{h}}^j_{u}\odot (\mathds{1}-\mathbf{e}_{u})\Vert^2_2\bigr], \\
		\end{split}
		\label{autoencoder_loss2}
	\end{equation}
	where $\mathbf{h}^j_{u}$ and $\overline{\mathbf{h}}^j_{u}$ are the $u$-th vector of $\mathbf{H}^j$ and $\overline{\mathbf{H}}^j$, respectively.
	$\mathds{1}$ is an indicator vector with the values of 1.
	
	Through the above operations, $\overline{\mathcal{E}}^j$ and $\overline{\mathbf{X}}^j$ are used to form the learnable potential graph $\overline{\mathcal{G}}^j=(\mathcal{V}^j, \overline{\mathcal{E}}^j, \overline{\mathbf{X}}^j)$.
	
	\textbf{Graph Fixing.}
	The edge server $S_j$ divides $\overline{\mathcal{G}}^j$ into some subgraphs, denoted by the set $\{\overline{\mathcal{G}}^{ji}(\mathcal{V}^{ji}, \overline{\mathcal{E}}^{ji}, \overline{\mathcal{N}}^{ji})|i\in [M_j]\}$,where $\overline{\mathcal{E}}^{ji}=\{\overline{e}^{ji}_{uv}|\overline{e}^{ji}_{uv} \in \overline{\mathcal{E}}^j, \forall u,v\in \mathcal{V}^{ji}\}$ is the neighbor set of $\mathcal{V}^{ji}$,  $\overline{\mathcal{N}}^{ji}=\{\overline{\mathcal{X}}^{ji}_u|u\in \mathcal{V}^{ji}\}$, and $\overline{\mathcal{X}}^{ji}_u=\{\overline{\mathbf{x}}^{ji}_v| \overline{e}^{ji}_{uv}\in\overline{\mathcal{E}}^{ji}\}$ indicates the potential neighbor feature vectors of $u$.
	Next, $S_j$ assigns the subgraphs to each client.
	It is noted that each local client repairs the subgraph by using the local graphic patcher $P_i^j$ referring to $\hat{\mathcal{G}}^{ji}=P^j_i(\overline{\mathcal{G}}^{ji})$.
	This process simulates the missing links, thereby promoting the feature propagation of local tasks in Eq. \eqref{graphsage}.
	By collaborating with the edge server, clients are expected to acquire diverse neighbor features from globally-shared information, thereby fixing cross-subgraph missing links.
	Moreover, these cross-subgraph links contribute to training a global node classifier $F_j$, aligning with the overall optimization objective in Eq. \eqref{loss1}.
	
	\subsection{SpreadFGL: Distributed Federated Graph Learning}
	In real-world application scenarios, a single edge server may encounter the problem of excessive costs and degraded performance as the number of clients expands, particularly when clients are geographically dispersed.
	To address this problem, we propose a novel distributed FGL framework, named SpreadFGL, that extends the FedGL to a multi-edge environment.
	The SpreadFGL is able to facilitate more efficient FGL training and better load balancing in a multi-edge collaborative environment.
	We consider that there are $N$ edge servers, and an edge server $S_j$ is equipped with a global node classifier $F_{j}$ parameterized by $\mathbf{W}_{j}$.
	Besides, a client only communicates with its closest edge server. There exist neighbor relationships among the servers, denoted by the matrix $\mathbf{A}\in \mathbb{R}^{N\times N}$.
	If $S_i$ and $S_j$ are neighbors, $a_{ij}=1$; otherwise, $a_{ij}=0$.
	Moreover, the parameter transmission is permitted between neighbor servers.
	
	In SpreadFGL, the clients adopt the $L$-layer GNNs and conduct the feature propagation via Eq. \eqref{graphsage} during the local training.
	The edge servers exchange information with the covered clients in each edge-client communication.
	At each $\mathcal{K}$ intervals of edge-client communications, the clients and their nearest edge servers collaboratively utilize the shared information to extract the potential links based on the proposed graph imputation generator and negative sampling mechanism.
	However, the potential cross-subgraph links strictly depend on the information provided by all clients.
	This not only violates the core idea of the SpreadFGL but also is impractical if the information is transmitted from the clients that are under the coverage of other servers.
	In light of these concerns, we design a weight regularizer during the local training.
	Based on trace normalization, the regularizer is used to enhance the network learning capability of the local node classifiers.
	Specifically, the loss function of the $i$-th client under the coverage of $S_j$ is defined as
	
	\vspace{-0.3cm}
	\begin{small}
		\begin{equation}
			\begin{split}
				\mathcal{L}_{F_i^j} &=\mathcal{R}^j_i\left(F^j_i(\mathbf{W}_{(j,i)})\right)\\
				&=-\sum_{u=1}^{|\mathcal{V}_t^{ji}|}\sum_{r=1}^c\mathbf{Y}_{ur}^{ji}ln\mathbf{H}^{(j,i)}
				+ \mbox{Tr}(\mathbf{W}_{(j,i,L)}\mathbf{W}_{(j,i,L)}^T), \\
			\end{split}
			\label{update_loss3}
		\end{equation}
	\end{small}
	\hspace{-0.15cm}where $\mbox{Tr}(\cdot)$ is the square matrix trace.
	$\mathbf{W}_{(j,i,L)}$ indicates the parameters of $L$-th GNN layer for the local node classifier $F^j_i$.
	
	To better explore the potential cross-subgraph links by using the information from other servers, we adopt the topology structure at the edge layer to facilitate the parameter transmission between neighbor servers.
	This enables the information flow among clients via the gradient propagation at each $\mathcal{K}$ intervals of edge-client communication.
	Specifically, $S_j$ first aggregates the model parameters of its neighbor servers.
	Next, $S_j$ averages the parameters and broadcast them to the covered clients. 
	This process can be described as
	\begin{equation}
		\begin{split}
			\mathbf{W}_{j}\leftarrow 1/({\displaystyle{\mathop{\sum}_{r=1}^{N}}a_{rj}M_r})\mathop{\sum}\limits_{r=1}\limits^{N}\mathop{\sum}\limits_{i=1}\limits^{M_r}a_{rj}\mathbf{W}_{(r,i)}. \\
		\end{split}
		\label{update_spread}
	\end{equation}
	
	The procedure of the proposed SpreadFGL is elaborated in Algorithm \ref{alg:alg1}, whose core components have been described in detail before.
	
	\begin{algorithm}[t]
		\caption{The proposed SpreadFGL.}
		\label{alg:alg1}
		\textbf{Input:} $N$ edge servers, $M$ clients, local graph datasets $\{\mathcal{D}^j_i\{\mathcal{G}^{ji}, \mathbf{Y}^{ji}\}| i \in [M_j]\}$, adjacency matrix $\mathbf{A}$, local training rounds $T_l$, edge-client communication rounds $T_g$, assessor iterations $T_{as}$, autoencoder iterations $T_{ae}$.
		
		\For{ $j=1 \rightarrow N$}
		{
			Initialize $\mathbf{W}_{(j,i)} \leftarrow \mathbf{W}_{j}$, $\forall i\in [M_j]$;
		}
		\For{$t_g=0\rightarrow T_g-1$}
		{
			\textbf{\# Parallel training of edge servers.} \\
			\For{$j=1\rightarrow N$}
			{
				\textbf{\# Training of clients.} \\
				\For {$t_l=0 \rightarrow T_l-1$}
				{
					Calculate $\mathcal{L}_{F_i^j}\hspace{-0.15cm}\leftarrow\hspace{-0.1cm}\mathcal{R}^j_{i}(F^j_i(\mathbf{W}_{(j,i)}))$ in Eq. \hspace{-0.15cm}\eqref{update_loss3};
				}
				\textbf{\# Training of graph imputation.} \\
				\eIf{$t_g \ \% \ \mathcal{K} = 0$}
				{
					Upload $S_j\leftarrow\{\mathbf{W}_{(j,i)}\}, \{\mathbf{H}^{(j,i)}\}, i\in[M_j]$; \\
					Aggregate $S_j\leftarrow\{\mathbf{W}_{(r,i)}|i\in [M_r]\}$, where $S_r$ is the neighbor of $S_j$; \\
					Update $\mathbf{W}_{j}$ in Eq. \eqref{update_spread}; \\
					Calculate $\overline{\mathbf{A}}^{j}\leftarrow \mathbf{H}^{j}{\mathbf{H}^{j}}^T$ and form links $\overline{\mathcal{E}}$;\\
					\While{\rm not convergent}
					{
						\For{$t_{ae}=0\rightarrow T_{ae}-1$}
						{
							Calculate {\small $\overline{\mathbf{H}}^{j}\leftarrow\hspace{-0.1cm} h(f(\mathbf{S}))$} by Eq. \eqref{autoencoder};\\
							Update $\Phi_{AE}^{t_{ae+1}}\hspace{-0.2cm}\leftarrow \hspace{-0.1cm}\Phi_{AE}^{t_{ae}}-\hspace{-0.05cm}\beta\nabla_{\Phi_{AE}^{t_{ae}}}\mathcal{L}_{AE}$ in Eq. \eqref{autoencoder_loss2};\\
						}
						\For{$t_{as}=0\rightarrow T_{as}-1$}
						{
							Calculate {\small $\overline{\mathbf{H}}^{j}\leftarrow\hspace{-0.1cm} h(f(\mathbf{S}))$} by Eq. \eqref{autoencoder};\\
							Update {\small $\Phi_{AS}^{t_{as+1}}\hspace{-0.2cm}\leftarrow\hspace{-0.1cm} \Phi_{AE}^{t_{as}}-\beta\nabla_{\Phi_{AS}^{t_{as}}}\mathcal{L}_{AS}$} in Eq. \eqref{assessor_loss2};\\
						} 
					}
					\textbf{\# Graph fixing in clients.} \\
					Download\hspace{-0.05cm} {\small $C^j_{i}\!\leftarrow\!\overline{\mathcal{G}}_{ji}\!(\mathcal{V}^{ji},\!\overline{\mathcal{E}}^{ji},\!\overline{\mathcal{N}}^{ji}),\forall i\in [M_j]$};\\
					Fix $\hat{\mathcal{G}}^i\leftarrow P_i(\overline{\mathcal{G}}^i)$, $\forall i\in [M_j]$;\\
				}{
					Aggregate $S_j\leftarrow\left\{\mathbf{W}_{(S_j,i)}|i\in [M_j]\right\}$; \\
					Calculate {\small $\mathbf{W}_{j}\leftarrow \frac{1}{M_j}\mathop{\sum}\limits_{i=1}\limits^{M_j}\mathbf{W}_{(j,i)}$};\\
				}
			} 
			Download $\mathbf{W}_{(j,i)}\leftarrow \mathbf{W}_{j}$, $\forall i\in [M_j]$; \\
		}
		\textbf{Output:} Global node classifiers $\left\{F_{j}|j\in [N]\right\}$.	
		\vspace{-0.1cm}
	\end{algorithm}

	\section{Performance Evaluation} \label{experiment}
	In this section, we first compare the proposed FedGL and SpreadFGL with state-of-the-art algorithms based on real-world testbed and graph datasets.
	Next, we conduct ablation experiments to further verify the superiority of the core components designed in the proposed frameworks.
	
	\begin{table*}[t]
		\centering
		\caption{Description of benchmark graph datasets}
		\resizebox{0.85\textwidth}{!}{
			\begin{tabular}{c|cccccccccccccccc}
				\toprule[1pt]
				\multicolumn{1}{c|}{Datasets}  & \multicolumn{4}{c}{Cora} & \multicolumn{4}{c}{Citeseer} & \multicolumn{4}{c}{WikiCS} & \multicolumn{4}{c}{CoauthorCS} \\ 
				\midrule
				\multicolumn{1}{c|}{c} & \multicolumn{4}{c}{7}     & \multicolumn{4}{c}{6}         & \multicolumn{4}{c}{10}       & \multicolumn{4}{c}{15}           \\
				\multicolumn{1}{c|}{$|\mathcal{V}|$}   & \multicolumn{4}{c}{2,708}     & \multicolumn{4}{c}{3,327}         & \multicolumn{4}{c}{11,701}       & \multicolumn{4}{c}{18,333}           \\
				\multicolumn{1}{c|}{$|\mathcal{E}|$}   & \multicolumn{4}{c}{5,429}     & \multicolumn{4}{c}{4,715}         & \multicolumn{4}{c}{215,863}       & \multicolumn{4}{c}{81,894}			\\
				\multicolumn{1}{c|}{$d$} & \multicolumn{4}{c}{1433}     & \multicolumn{4}{c}{3703}         & \multicolumn{4}{c}{300}       & \multicolumn{4}{c}{6,805}           \\
				\midrule
				$M$                        & 6    & 9   & 12   & 15   & 6     & 9    & 12    & 15    & 6    & 9    & 12    & 15   & 6     & 9     & 12     & 15    \\ 
				\cmidrule{1-1}\cmidrule(r){2-5} \cmidrule(lr){6-9} \cmidrule(lr){10-13} \cmidrule(l){14-17}
				$|\overline{\mathcal{V}_i}|$                         &451      &300     &225      &180      &554      &369      &277       &221       &1,950      &1,300      &975       &780      &3,055       &2,037       &1,527        &1,222       \\
				$|\overline{\mathcal{E}_i}|$                         &750      &438     &304      &229      &768      &357      &341       &263       &13,928      &6,607      &3,985       &2,685      &7,582       &4,047       &2,475        &1,947       \\
				$|\Delta\mathcal{E}|$                   &935      &1,487     &1,781     &1,994      &110       &434      &632       &782      &348,157      &372,257      &383,899       &291,439      &36,405       &45,475       &52,203        &52,699       \\
				\bottomrule[1pt]   
		\end{tabular}}
		\label{dataset}
		\vspace{-0.5cm}
	\end{table*}

	\subsection{Experiment Setup}
	\textbf{Real-world Testbed.}
	As shown in Fig. 3, we build a hardware testbed to evaluate the proposed FedGL and SpreadFGL in real-world scenarios of edge-client collaboration.
	In the testbed, each Raspberry Pi 4B acts as a client that is equipped with Broadcom BCM2711 SoC @1.5GHz with 4 GB RAM, and the OS is Raspbian GNU/Linux 11.
	Each Jetson TX2  acts as an edge server that is equipped with one 256-core NVIDIA Pascal(R) GPU, one Dual-core Denver 2 64-bit CPU and a quad-core Arm(R) Cortex(R)-A57 MPCore processor equipped with 8 GB RAM, and the OS is Ubuntu 18.04.6 LTS.
	The above hardware communicates through a 5 GHz WiFi network, and the proposed frameworks are implemented based on PyTorch.
	After completing local training, the client (Raspberry Pi 4B) uploads the local model parameters to its connected edge server (Jetson TX2).
	An edge server conducts aggregation and distributes the globally-shared model to its connected clients.
	
	\begin{figure}[t]
		\centering
		\begin{center}
			\includegraphics*[width=0.68\linewidth]{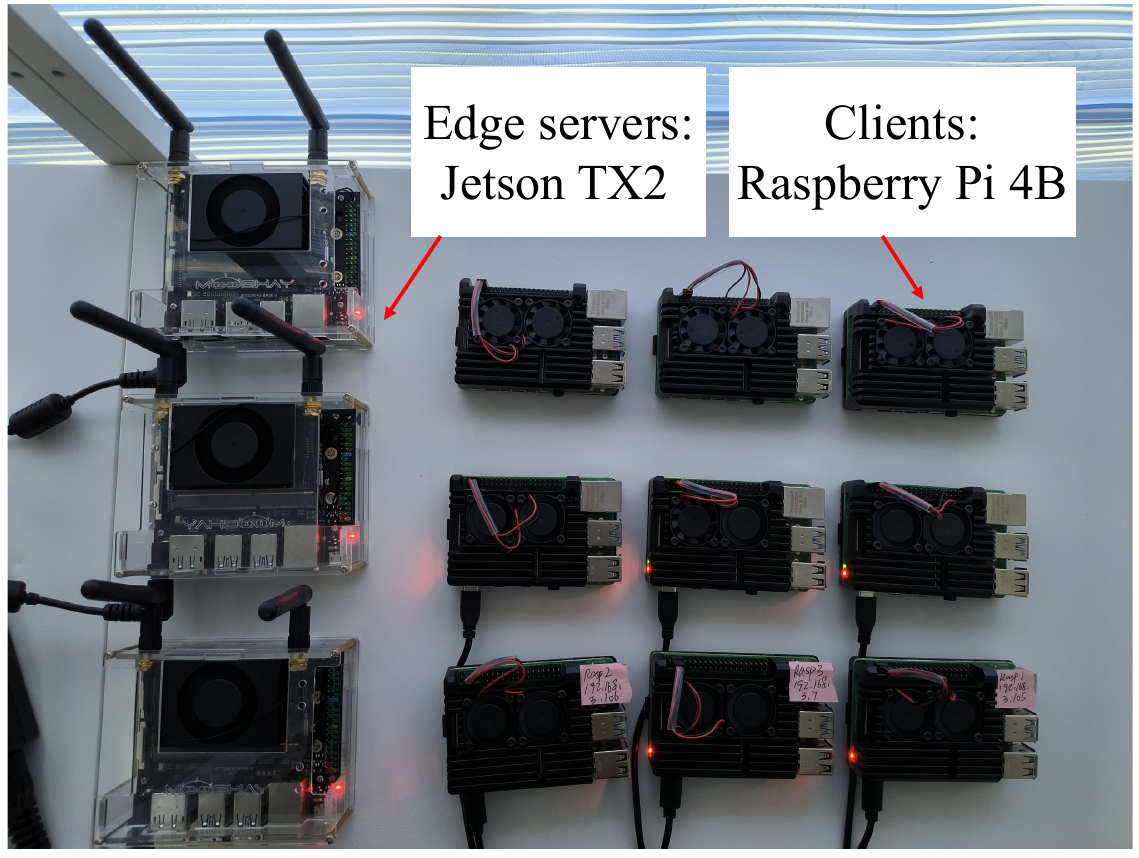}
			\vspace{-0.3cm}
			\caption{Real-world testbed for FedGL and SpreadFG.}
			\label{pi_setting}
			\vspace{-0.7cm}
		\end{center}
	\end{figure}
	
	\textbf{Datasets.}
	The following four benchmark graph datasets are used in our experiments, as shown in Table \ref{dataset}, where $c$ is the number of classes, $|\overline{V_i}|$ and $|\overline{\mathcal{E}_i}|$ are the average number of nodes and edges in subgraphs, and $|\Delta \mathcal{E}|$ is the number of missing cross-subgraph links.
	\begin{itemize}
		\item \textbf{Cora} \cite{sen2008collective} is a dataset of citation network, where nodes and edges indicate papers and their mutual citations, respectively.
		According to the paper topics, the nodes are labeled with $7$ classes.
		\item \textbf{Citeseer} \cite{sen2008collective} is a research paper citation dataset, where nodes and edges indicate publications and citation relationships, respectively.
		The citation relationships are defined as a word vector, and the nodes are classified into $6$ classes.
		\item \textbf{WikiCS} \cite{mernyei2020wiki} is a dataset derived from Wikipedia, where nodes and edges indicate computer science (CS) articles and different branches, respectively.
		All nodes are labeled with $6$ classes.
		\item \textbf{CoauthorCS} \cite{shchur2018pitfalls} is an academic network dataset on microsoft scholar graph, where nodes and edges indicate authors and co-author relationships, respectively.
		The nodes are labeled with $15$ classes based on research fields.
	\end{itemize}
	
	\textbf{Comparison Algorithms.}
	We compare our proposed FedGL and SpreadFGL with the following state-of-the-art algorithms.
	\begin{itemize}
		\item \textbf{LocalFGL} is a local node classifier in the SpreadFGL, which is trained by a client independently.
		\item \textbf{FedAvg-fusion} \cite{DBLP:conf/aistats/McMahanMRHA17} is an improved FedAvg framework, which trains a globally-shared GNN model with FedAvg via collaborating subgraphs distributed among clients.
		\item \textbf{FedSage+} \cite{zhang2021subgraph} adopts a linear predictor to locally repair the potential links between subgraphs, referring to the latent information in each training round.
	\end{itemize}
	
	It is worth noting that there are few studies for handling the FGL scenario with completely missing cross-subgraph links between clients.
	FedSage+ is deemed as the state-of-the-art algorithm for studying the missing cross-subgraph links in FGL fields.
	However, it still suffers from performance bottlenecks and has not been well solved in real-world scenarios.
	
	\textbf{Parameter Settings.}
	For the proposed SpreadFGL and FedGL, we adopt the GraphSAGE \cite{DBLP:conf/nips/HamiltonYL17} with two layers and use the GCN aggregator as local node classifiers. 
	The autoencoder employs $4$ fully-connected layers, where the neural number of encoder and decoder are $\{c,16,d\}$ and $\{d,16,c\}$, respectively.
	In the autoencoder, the Softmax is used as an activation function in the last layer.
	The assessor adopts a fully-connected neural network, where the hidden neural number is $\{c,128,16,1\}$.
	In the assessor, the Sigmoid is used as an activation function in the last layer while the ReLU is used in the rest layers.
	The training iterations of the autoencoder and assessor are $T_{ae}=5$ and $T_{as}=3$, respectively, and the Adam optimizer is used to update parameters with the learning rate of $0.001$.
	The threshold $\theta$ is set to $1/c$ and $k$ ranges in $[3,20]$.
	Moreover, we select $[20\%, 60\%]$ samples as the training set and randomly choose $20\%$ as the testing set.
	The Louvain algorithm \cite{blondel2008fast} is used to measure the subgraph similarity for clients.
	The FedGL uses an edge server and the SpreadFGL adopts three edge servers for collaborative training with a ring topology structure, where the number of clients ranges in $[6,15]$.
	The Adam optimizer is used to update the parameters of local classifiers with the learning rate $lr=0.01$.
	Besides, we use the well-known accuracy (ACC) and macro F1-score (F1) as performance metrics.
	
	\subsection{Experiment Results and Analysis}
	\textbf{Node Classification Accuracy.}
	As shown in Table \ref{result_1}, the proposed SpreadFGL and FedGL can both achieve higher classification accuracy than other state-of-the-art algorithms under different datasets, indicating the superiority of the proposed frameworks for node classification tasks.
	Specifically, the significant performance gap between the LocalFGL and SpreadFGL verifies the advantages of using the proposed edge-client collaboration mechanism.
	The FedGL and SpreadFGL outperform the FedSage+ by around $12.78\%$ and $14.71\%$ in terms of ACC and F1, respectively.
	This demonstrates that the FedGL and SpreadFGL gain more generalized potential cross-subgraph links through the global information flow, further validating the effectiveness of the proposed graph imputation generator.
	Moreover, compared to the FedGL, the SpreadFGL achieves better performance on most of the datasets under various scenarios with different numbers of clients.
	This indicates that the information flow between clients and edge servers utilized in the SpreadFGL effectively promotes the repair of missing links among clients even though the scenario becomes complex with more clients.
	
	\begin{table*}[t]
		\centering
		\caption{Node classification accuracy (\%) on four datasets with labeled ratio of $0.3$ and $M=6,9,12,16$}
		\vspace{-0.2cm}
		\resizebox{0.7\textwidth}{!}{
			\begin{tabular}{cccccccccc}
				\toprule[1pt]
				\multicolumn{2}{c}{Dataset}                                                     & \multicolumn{4}{c}{Cora}                                                                        & \multicolumn{4}{c}{Citeseer}                                                                    \\ 
				\cmidrule{1-2}\cmidrule(lr){3-6} \cmidrule(lr){7-10}
				Methods                                         & \multicolumn{1}{l}{Metrics} & \multicolumn{1}{l}{$M=6$} & \multicolumn{1}{l}{$M=9$} & \multicolumn{1}{l}{$M=12$} & \multicolumn{1}{l}{$M=15$} & \multicolumn{1}{l}{$M=6$} & \multicolumn{1}{l}{$M=9$} & \multicolumn{1}{l}{$M=12$} & \multicolumn{1}{l}{$M=15$} \\ 
				\midrule[1pt]
				\multicolumn{1}{c||}{\multirow{2}{*}{LocalFGL}}  & \multicolumn{1}{c||}{ACC}      &62.20                       &60.00                       &57.14                        &63.33                        &51.63                       &55.56                       &46.15                        &43.75                        \\
				\multicolumn{1}{c||}{}                           & \multicolumn{1}{c||}{F1}       &56.71                       &52.43                       &53.96                        &41.47                        &47.85                       &49.70                       &46.00                        &37.90                        \\ 
				\cmidrule(r){1-6} \cmidrule(l){7-10}
				\multicolumn{1}{c||}{\multirow{2}{*}{FedAvg-fusion}}    & \multicolumn{1}{c||}{ACC}     &\underline{81.70}                      &76.89                     &73.19                        &\underline{70.61}                        &71.57                       &71.42                       &69.07                        &68.64                        \\
				\multicolumn{1}{c||}{}                           & \multicolumn{1}{c||}{F1}       & 79.15                     &74.05                       &66.14                        &\underline{63.83}                        &61.89                       &\underline{67.17}                      &60.00                        &60.11                        \\ 
				\cmidrule(r){1-6} \cmidrule(l){7-10}
				\cmidrule(r){1-6} \cmidrule(l){7-10}
				\multicolumn{1}{c||}{\multirow{2}{*}{FedSage+}}  & \multicolumn{1}{c||}{ACC}      &80.26                       &\underline{80.18}                      &\underline{80.06}                        &48.11                        &\underline{73.13}                      &\underline{72.87}                       &\underline{72.46}                        &\underline{72.09}                        \\
				\multicolumn{1}{c||}{}                           & \multicolumn{1}{c||}{F1}       &\underline{79.98}                       &\underline{79.63}                       &\underline{78.72}                        &48.06                        &\underline{63.12}                       &62.25                       &\underline{61.65}                        &\underline{60.45}                       \\ 
				\midrule[1pt]
				\multicolumn{1}{c||}{\multirow{2}{*}{FedGL}}    & \multicolumn{1}{c||}{ACC}      &\textcolor{blue}{\textbf{84.47}}                      &\textcolor{blue}{\textbf{83.36}}                       &\textcolor{red}{\textbf{82.81}}                        &\textcolor{blue}{\textbf{76.71}}                        &\textcolor{red}{\textbf{73.83}}                       &\textcolor{blue}{\textbf{73.08}}                       &\textcolor{blue}{\textbf{73.53}}                        &\textcolor{blue}{\textbf{73.03}}                        \\
				\multicolumn{1}{c||}{}                           & \multicolumn{1}{c||}{F1}       &\textcolor{blue}{\textbf{84.08}}                       &\textcolor{blue}{\textbf{83.11}}                       &\textcolor{blue}{\textbf{81.63}}                       &\textcolor{blue}{\textbf{75.34}}                        &\textcolor{red}{\textbf{69.41}}                      &\textcolor{blue}{\textbf{67.53}}                       &\textcolor{blue}{\textbf{64.39}}                       &\textcolor{blue}{\textbf{63.72} }                     \\ 
				\cmidrule(r){1-6} \cmidrule(l){7-10}
				\multicolumn{1}{c||}{\multirow{2}{*}{SpreadFGL}} & \multicolumn{1}{c||}{ACC}      &\textcolor{red}{\textbf{84.49}}                       &\textcolor{red}{\textbf{83.56}}                       &\textcolor{blue}{\textbf{82.59}}                        &\textcolor{red}{\textbf{78.55}}                        &\textcolor{blue}{\textbf{73.38}}                       &\textcolor{red}{\textbf{73.43}}                       &\textcolor{red}{\textbf{73.72}}                       &\textcolor{red}{\textbf{73.23}}                        \\
				\multicolumn{1}{c||}{}                           & \multicolumn{1}{c||}{F1}       &\textcolor{red}{\textbf{84.32}}                       &\textcolor{red}{\textbf{83.11}}                       &\textcolor{red}{\textbf{82.34}}                        &\textcolor{red}{\textbf{75.90}}                        &\textcolor{blue}{\textbf{67.72}}                       &\textcolor{red}{\textbf{68.12}}                       &\textcolor{red}{\textbf{67.01}}                        &\textcolor{red}{\textbf{67.63}}                       \\ 
				\midrule[1pt]
				\multicolumn{2}{c}{Dataset}                                                     & \multicolumn{4}{c}{WikiCS}                                                                        & \multicolumn{4}{c}{CoauthorCS}                                                                    \\ 
				\cmidrule{1-2}\cmidrule(lr){3-6} \cmidrule(lr){7-10}
				Methods                                         & \multicolumn{1}{l}{Metrics} & \multicolumn{1}{l}{$M=6$} & \multicolumn{1}{l}{$M=9$} & \multicolumn{1}{l}{$M=12$} & \multicolumn{1}{l}{$M=15$} & \multicolumn{1}{l}{$M=6$} & \multicolumn{1}{l}{$M=9$} & \multicolumn{1}{l}{$M=12$} & \multicolumn{1}{l}{$M=15$} \\ 
				\midrule[1pt]
				\multicolumn{1}{c||}{\multirow{2}{*}{LocalFGL}}  & \multicolumn{1}{c||}{ACC}      &58.58                       &55.56                       &52.13                        &47.46                        &82.76                       &80.00                       &79.81                        &79.90                        \\
				\multicolumn{1}{c||}{}                           & \multicolumn{1}{c||}{F1}       &52.06                       &48.50                       &46.06                        &42.31                        &57.45                       &58.55                       &53.97                        &62.06                        \\ 
				\cmidrule(r){1-6} \cmidrule(l){7-10}
				\multicolumn{1}{c||}{\multirow{2}{*}{FedAvg-fusion}}    & \multicolumn{1}{c||}{ACC}     &\underline{76.25}                       &\underline{74.70}                       &\underline{73.67}                        &\underline{73.37}                        &\underline{87.73}                       &86.96                       &87.35                        &87.60                        \\
				\multicolumn{1}{c||}{}                           & \multicolumn{1}{c||}{F1}       &\underline{68.98}                       &\underline{66.52}                       &\underline{63.09}                        &\underline{62.53}                        &\underline{73.46}                       &\underline{67.15}                       &\underline{62.68}                        &64.11                        \\ 
				\cmidrule(r){1-6} \cmidrule(l){7-10}
				\cmidrule(r){1-6} \cmidrule(l){7-10}
				\multicolumn{1}{c||}{\multirow{2}{*}{FedSage+}}  & \multicolumn{1}{c||}{ACC}      &36.32                       &38.73                       &36.94                        &38.89                        &86.93                       &\underline{87.69}                       &\underline{87.68}                        &\underline{88.03}                        \\
				\multicolumn{1}{c||}{}                           & \multicolumn{1}{c||}{F1}       &32.32                       &34.56                       &33.24                        &35.61                        &66.57                       &67.06                       &61.85                        &\underline{65.06}                        \\ 
				\midrule[1pt]
				\multicolumn{1}{c||}{\multirow{2}{*}{FedGL}}    & \multicolumn{1}{c||}{ACC}      &\textcolor{blue}{\textbf{77.56}}                       &\textcolor{blue}{\textbf{76.97}}                       &\textcolor{blue}{\textbf{76.24}}                        &\textcolor{blue}{\textbf{75.26}}                        &\textcolor{red}{\textbf{90.49}}                    &\textcolor{blue}{\textbf{89.74}}                     &\textcolor{blue}{\textbf{87.72}}                      &\textcolor{blue}{\textbf{88.62}}                      \\
				\multicolumn{1}{c||}{}                           & \multicolumn{1}{c||}{F1}       &\textcolor{blue}{\textbf{70.71}}                       &\textcolor{blue}{\textbf{68.87}}                       &\textcolor{blue}{\textbf{66.83}}                        &\textcolor{blue}{\textbf{64.37}}                        &\textcolor{red}{\textbf{74.89}}                       &\textcolor{blue}{\textbf{67.38} }                      &\textcolor{blue}{\textbf{65.22}}                        &\textcolor{blue}{\textbf{65.06}}                        \\ 
				\cmidrule(r){1-6} \cmidrule(l){7-10}
				\multicolumn{1}{c||}{\multirow{2}{*}{SpreadFGL}} & \multicolumn{1}{c||}{ACC}      &\textcolor{red}{\textbf{78.93}}                       &\textcolor{red}{\textbf{78.06}}                       &\textcolor{red}{\textbf{77.10}}                        &\textcolor{red}{\textbf{76.32}}                        &\textcolor{blue}{\textbf{90.43}}                       &\textcolor{red}{\textbf{89.74}}                       &\textcolor{red}{\textbf{89.68}}                        &\textcolor{red}{\textbf{88.63}}                        \\
				\multicolumn{1}{c||}{}                           & \multicolumn{1}{c||}{F1}       &\textcolor{red}{\textbf{72.19}}                       &\textcolor{red}{\textbf{71.32}}                       &\textcolor{red}{\textbf{69.78}}                        &\textcolor{red}{\textbf{67.49}}                       &\textcolor{blue}{\textbf{74.54}}                       &\textcolor{red}{\textbf{68.13}}                       &\textcolor{red}{\textbf{65.25}}                        &\textcolor{red}{\textbf{65.98}}                        \\ 
				\bottomrule[1pt]
		\end{tabular}}
		\label{result_1}
		\vspace{-0.3cm}
	\end{table*}
	
	\begin{figure}[t]
		\centering
		\begin{center}
			\includegraphics*[width=0.94\linewidth]{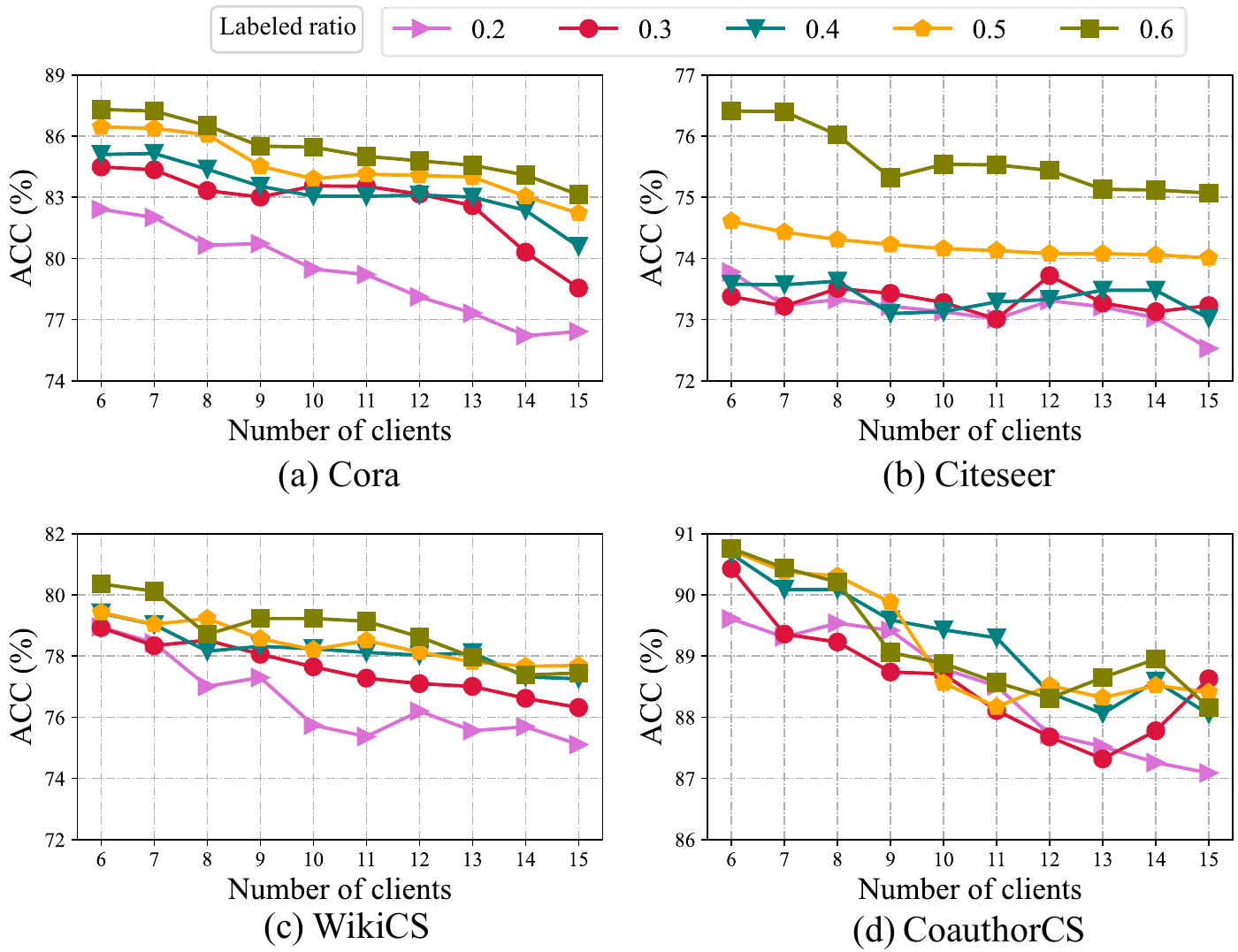}
			\vspace{-0.3cm}
			\caption{ACC of SpreadFGL with various numbers of clients and labeled ratios.
			}
			\label{different_ratio}
			\vspace{-0.4cm}
		\end{center}
	\end{figure}

	\textbf{Performance with Different Labeled Ratios.}
	Fig. \ref{different_ratio} depicts the ACC of the SpreadFGL on different datasets with various labeled ratios, varying from $0.2$ to $0.6$.
	With the same labeled ratio, the ACC tends to decrease as the datasets are distributed on more clients. 
	This is because massive heterogeneous clients cause difficulty and instability in the aggregation process of model parameters.
	Under this scenario, the performance of the classic FGL might be seriously degraded since it adopts a centralized training manner.
	It is noted that the ACC is rising as the labeled ratio increases, but with fewer data points presenting the opposite situation.
	This discrepancy may be attributed to the sparsity of certain classes in the feature space, leading to insufficient model training and thus affecting classification accuracy.
	
	\begin{figure}[t]
		\centering
		\begin{center}
			\includegraphics*[width=0.88\linewidth]{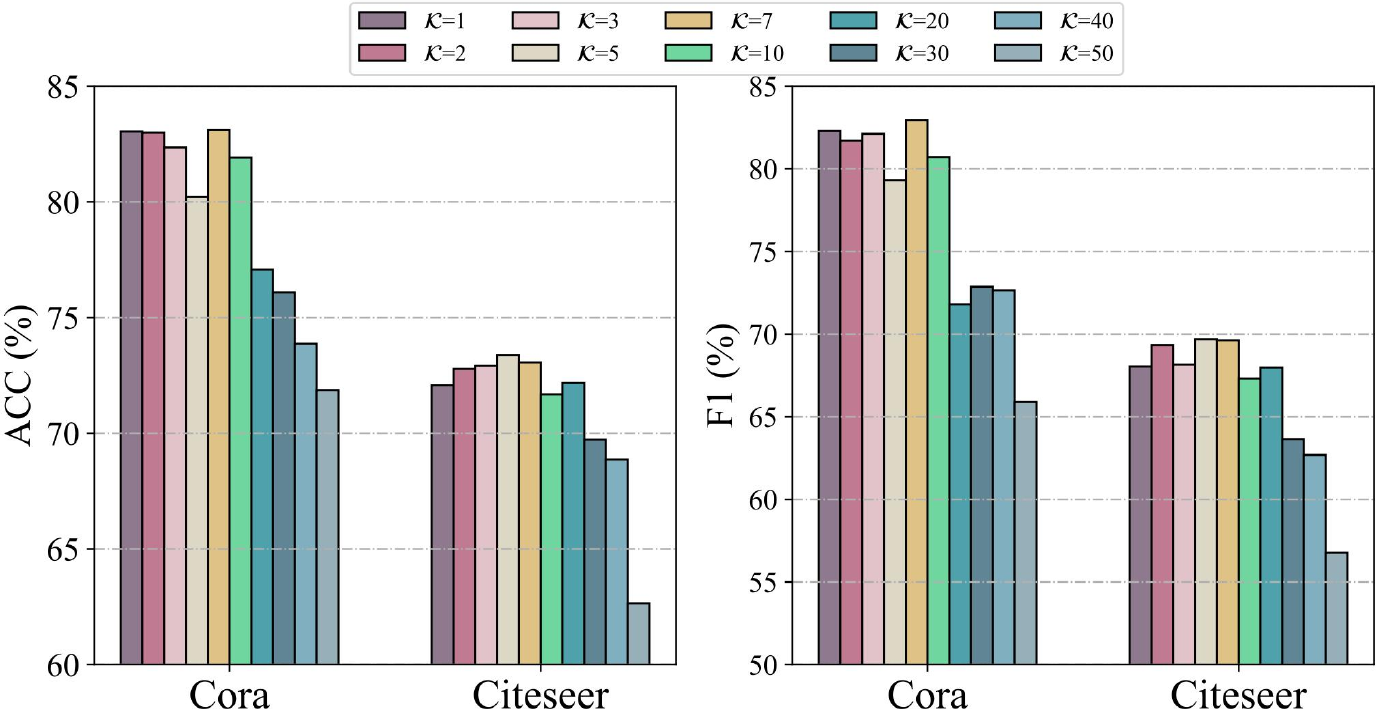}
			\vspace{-0.2cm}
			\caption{Accuracy of SpreadFGL with different values of $\mathcal{K}$.
			}
			\label{k_sen}
			\vspace{-0.4cm}
		\end{center}
	\end{figure}
	
	\begin{figure}[t]
		\centering
		\begin{center}
			\includegraphics*[width=0.88\linewidth]{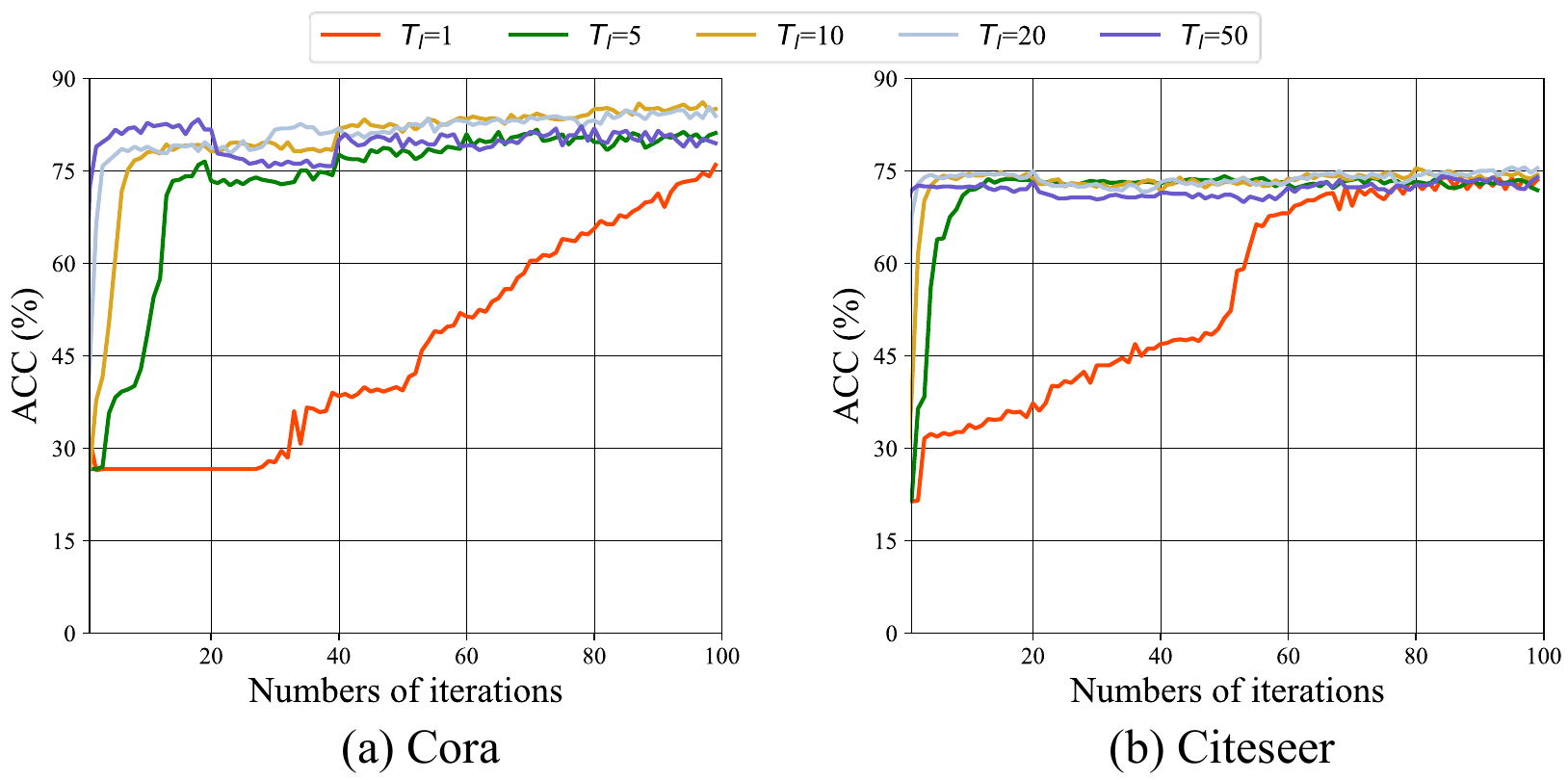}
			\caption{ACC of SpreadFGL with various local training iterations $T_l$.}
			\label{different_local_training}
			\vspace{-0.3cm}
		\end{center}
	\end{figure}
	\textbf{Parameter Sensitivity.}
	We analyze the parameter sensitivity of the proposed SpreadFGL on different datasets with respect to the hyperparameter $\mathcal{K}$ and $T_l$.
	As shown in Fig. \ref{k_sen}, $\mathcal{K}$ remarkably affects the classification accuracy in terms of the ACC and F1.
	Specifically, the ACC and F1 stay at a low level when $\mathcal{K}$ is more than $10$, while they keep stable as $\mathcal{K}$ ranges in $[1,10]$, attributed to the reason that the graph imputation generator can better repair the missing links in subgraphs to promote feature propagation in local models within fewer edge-client communications, thereby improving the training of the global node classifiers.
	In this regard, the suggested values of $\mathcal{K}$ range from $1$ to $10$.
	Fig. \ref{different_local_training} presents the influence of local training iteration $T_l$ on the SpreadFGL.
	The SpreadFGL converges slowly and achieves a local optimum when $T_l$ is less than $5$.
	This is because local models cannot sufficiently learn feature patterns within fewer local iterations, leading to slow model convergence.
	It is noted that the ACC declines when $T_l$ exceeds $50$ due to the overfitting of the model. 
	Therefore, a suitable range of $T_l$ is $[10,20]$, considering both accuracy and convergence speed.
	
	\begin{figure}[t]
		\centering
		\begin{center}
			\includegraphics*[width=0.94\linewidth]{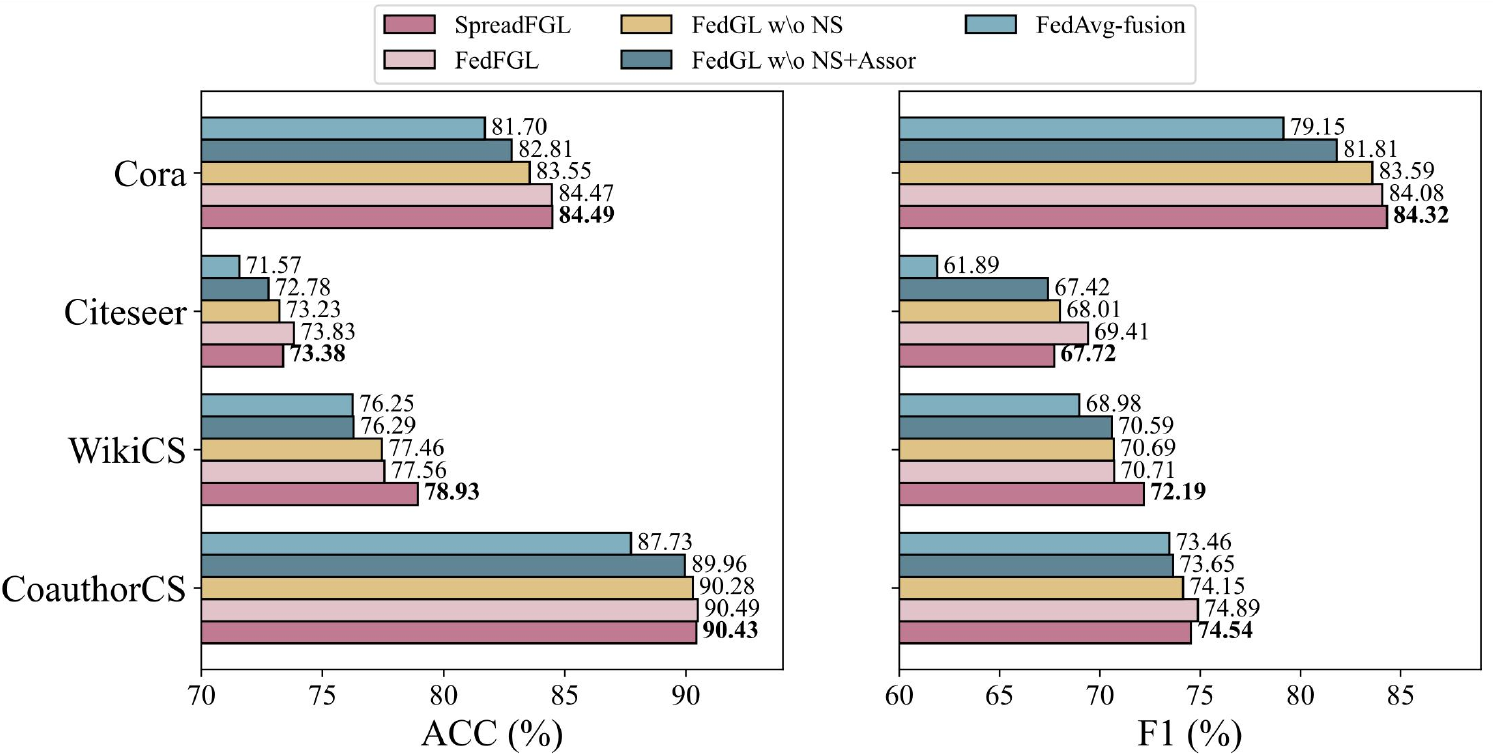}
			\vspace{-0.2cm}
			\caption{Ablation study on negative sampling mechanism and versatile assessor when $M=6$ and the labeled ratio is $0.3$.
			}
			\label{ablation}
			\vspace{-0.4cm}
		\end{center}
	\end{figure}
	
	\textbf{Ablation Study.}
	As shown in Fig. \ref{ablation}, we regard the FedAvg-fusion as a baseline that adopts the FedAvg to aggregate the parameters from multiple clients on an edge server.
	Also, we test the performance of the FedGL without a negative sampling mechanism (denoted by NS), versatile assessor (denoted by Assor), and the FedGL without NS.
	The proposed FedGL and SpreadFGL achieve comparable performance and outperform others by combining graph imputation generator, versatile assessor, and negative sampling mechanism.
	It is noted that there is only a small performance improvement when just utilizing one of the core components designed in the proposed frameworks.
	It obtains considerable improvement when the SpreadFGL adopts all the proposed components.
	This demonstrates that the integration of these components is able to better extract more refined potential cross-subgraph links, thereby promoting the accuracy of classification tasks.
	\begin{figure}[h]
		\centering
		\begin{center}
			\includegraphics*[width=0.94\linewidth]{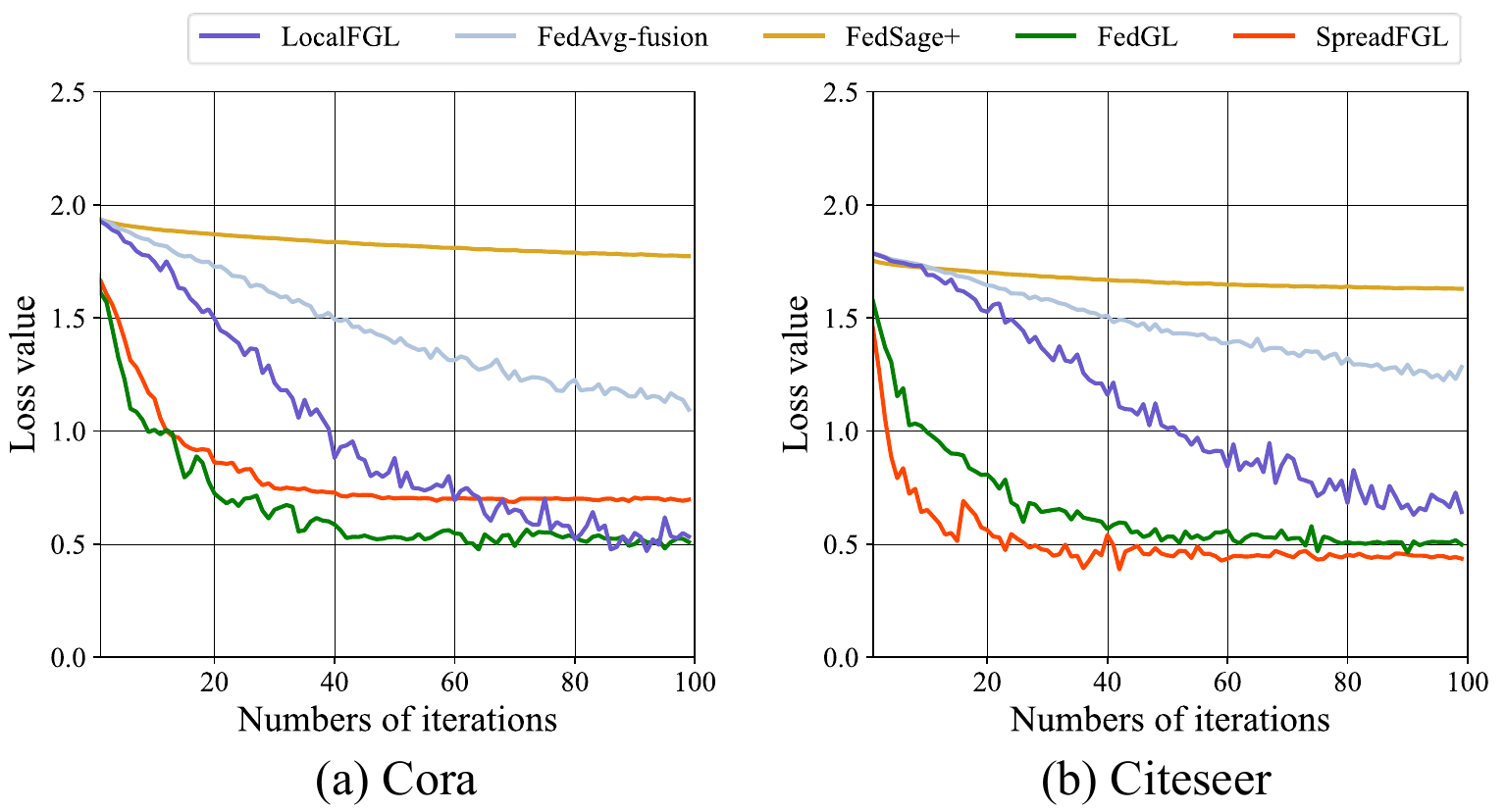}
			\vspace{-0.2cm}
			\caption{Training loss of different FGL-based frameworks when $M=6$ and the labeled ratio is $0.3$.
			}
			\label{loss_convergence}
		\end{center}
		\vspace{-0.4cm}
	\end{figure}
	
\begin{figure}[h]
		\centering
		\vspace{-0.2cm}
		\begin{center}
			\includegraphics*[width=0.94\linewidth]{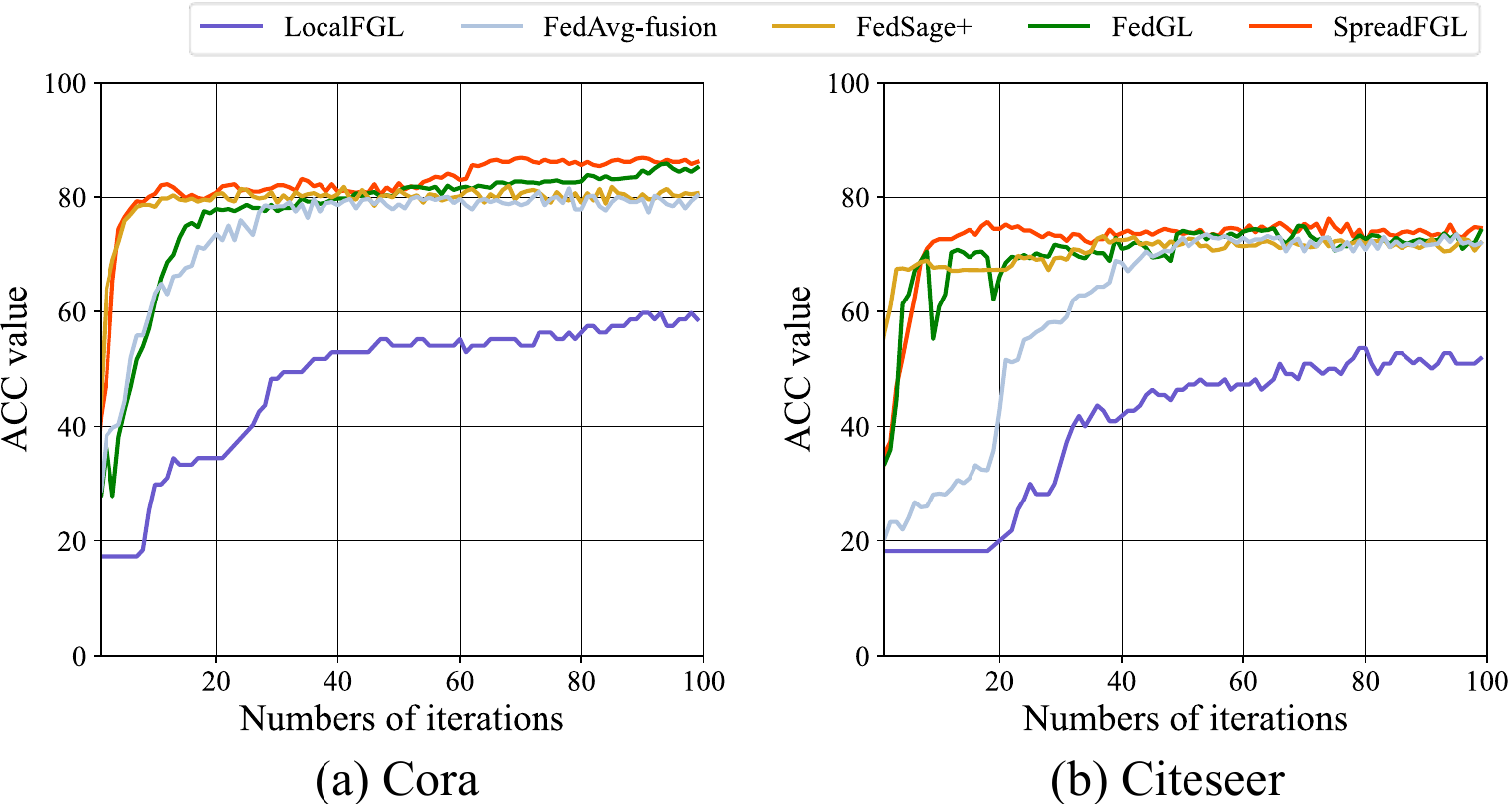}
			\vspace{-0.2cm}
			\caption{ACC of different FGL-based frameworks when $M=6$ and the labeled ratio is $0.3$.
			}
			\label{acc_convergence}
			\vspace{-0.5cm}
		\end{center}
	\end{figure}

	\textbf{Convergence Validation.}
	Fig. \ref{loss_convergence} illustrates the training loss of different FGL-based frameworks on Cora and Citeseer datasets.
	It can be observed that both the FedGL and SpreadFGL can always rapidly converge compared to the state-of-the-art algorithms, validating the effectiveness of the proposed frameworks in node classification tasks.
	Fig. \ref{acc_convergence} shows the curves of ACC when using different FGL-based frameworks.
	It is noted that the FedGL and SpreadFGL can achieve higher ACC than other state-of-the-art algorithms within fewer training iterations.
	Compared to the FedGL, the SpreadFGL converges faster to higher accuracy on different graph datasets, demonstrating the superiority of the SpreadFGL in multi-edge collaborative environments. 
	
	\section{Conclusion} \label{conclusion}
	In this paper, we propose a novel FGL-based framework named FedGL and its extended framework SpreadFGL, addressing the challenges of generating cross-subgraph links and single-node overloading.
	First, we design the FedGL to repair the missing links between clients, where a new graph imputation generator is developed that incorporates a versatile assessor and negative sampling mechanism to explore refined global information flow, extracting unbiased latent links and thus improving the training effect.
	Next, to alleviate the overloading issue at the edge layer, we extend the FedGL and propose the SpreadFGL with multi-edge collaboration to enhance the global information exchange.
	Extensive experiments are conducted on real-world testbed and benchmark graph datasets to verify the superiority of the proposed FedGL and SpreadFGL.
	The results show that the FedGL and SpreadFGL outperform state-of-the-art algorithms in terms of model accuracy.
	Further, through ablation experiments and convergence analysis, we validate the effectiveness of the core components designed in the proposed frameworks and the advantage of the SpreadFGL for achieving faster convergence speed.
	
	\section*{Acknowledgement}
	This work was partly supported by the National Natural Science Foundation of China (Grant No. 62202103), the Central Funds Guiding the Local Science and Technology Development (Grant No. 2022L3004), the Fujian Province Technology and Economy Integration Service Platform (Grant No. 2023XRH001), and the Fuzhou-Xiamen-Quanzhou National Independent Innovation Demonstration Zone Collaborative Innovation Platform (Grant No. 2022FX5).
	
	\bibliographystyle{ieeetr}
	\bibliography{SpreadFGL}

\begin{thebibliography}{10}

\bibitem{ling2022malgraph}
X.~Ling, L.~Wu, W.~Deng, Z.~Qu, J.~Zhang, S.~Zhang, T.~Ma, B.~Wang, C.~Wu, and
  S.~Ji, ``Malgraph: Hierarchical graph neural networks for robust windows
  malware detection,'' in {\em IEEE Conference on Computer Communications
  (INFOCOM)}, pp.~1998--2007, IEEE, 2022.

\bibitem{xie2021federated}
H.~Xie, J.~Ma, L.~Xiong, and C.~Yang, ``Federated graph classification over
  non-iid graphs,'' in {\em Advances in Neural Information Processing Systems
  (NIPS)}, pp.~18839--18852, 2021.

\bibitem{peng2021differentially}
H.~Peng, H.~Li, Y.~Song, V.~Zheng, and J.~Li, ``Differentially private
  federated knowledge graphs embedding,'' in {\em Proceedings of the ACM
  International Conference on Information and Knowledge Management (CIKM)},
  pp.~1416--1425, 2021.

\bibitem{chen2022graph}
J.~Chen, G.~Huang, H.~Zheng, S.~Yu, W.~Jiang, and C.~Cui, ``Graph-fraudster:
  Adversarial attacks on graph neural network-based vertical federated
  learning,'' {\em IEEE Transactions on Computational Social Systems (TCSS)},
  vol.~10, no.~2, pp.~492--506, 2022.

\bibitem{geyer2019deeptma}
F.~Geyer and S.~Bondorf, ``Deeptma: Predicting effective contention models for
  network calculus using graph neural networks,'' in {\em IEEE Conference on
  Computer Communications (INFOCOM)}, pp.~1009--1017, IEEE, 2019.

\bibitem{han2021tailored}
Y.~Han, S.~Shen, X.~Wang, S.~Wang, and V.~C. Leung, ``Tailored learning-based
  scheduling for kubernetes-oriented edge-cloud system,'' in {\em IEEE
  Conference on Computer Communications (INFOCOM)}, pp.~1--10, IEEE, 2021.

\bibitem{scardapane2020distributed}
S.~Scardapane, I.~Spinelli, and P.~Di~Lorenzo, ``Distributed training of graph
  convolutional networks,'' {\em IEEE Transactions on Signal and Information
  Processing over Networks (TSIPN)}, vol.~7, pp.~87--100, 2020.

\bibitem{zhang2021subgraph}
K.~Zhang, C.~Yang, X.~Li, L.~Sun, and S.~M. Yiu, ``Subgraph federated learning
  with missing neighbor generation,'' in {\em Advances in Neural Information
  Processing Systems (NIPS)}, pp.~6671--6682, 2021.

\bibitem{li2021sample}
A.~Li, L.~Zhang, J.~Tan, Y.~Qin, J.~Wang, and X.-Y. Li, ``Sample-level data
  selection for federated learning,'' in {\em IEEE Conference on Computer
  Communications (INFOCOM)}, pp.~1--10, IEEE, 2021.

\bibitem{zhang2021fastgnn}
C.~Zhang, S.~Zhang, J.~James, and S.~Yu, ``Fastgnn: A topological information
  protected federated learning approach for traffic speed forecasting,'' {\em
  IEEE Transactions on Industrial Informatics (TII)}, vol.~17, no.~12,
  pp.~8464--8474, 2021.

\bibitem{chen2021fedgraph}
F.~Chen, P.~Li, T.~Miyazaki, and C.~Wu, ``Fedgraph: Federated graph learning
  with intelligent sampling,'' {\em IEEE Transactions on Parallel and
  Distributed Systems (TPDS)}, vol.~33, no.~8, pp.~1775--1786, 2021.

\bibitem{wang2023graph}
Z.~Wang, Y.~Zhou, Y.~Zou, Q.~An, Y.~Shi, and M.~Bennis, ``A graph neural
  network learning approach to optimize ris-assisted federated learning,'' {\em
  IEEE Transactions on Wireless Communications (TWC)}, 2023.
\newblock doi: {10.1109/TWC.2023.3239400}.

\bibitem{wang2022graphfl}
B.~Wang, A.~Li, M.~Pang, H.~Li, and Y.~Chen, ``Graphfl: A federated learning
  framework for semi-supervised node classification on graphs,'' in {\em IEEE
  International Conference on Data Mining (ICDM)}, pp.~498--507, IEEE, 2022.

\bibitem{DBLP:conf/aaai/0001CBAA22}
C.~He, E.~Ceyani, K.~Balasubramanian, M.~Annavaram, and S.~Avestimehr,
  ``Spreadgnn: Decentralized multi-task federated learning for graph neural
  networks on molecular data,'' in {\em Proceeding of the {AAAI} Conference on
  Artificial Intelligence}, pp.~6865--6873, 2022.

\bibitem{yuan2022fedstn}
X.~Yuan, J.~Chen, J.~Yang, N.~Zhang, T.~Yang, T.~Han, and A.~Taherkordi,
  ``Fedstn: Graph representation driven federated learning for edge computing
  enabled urban traffic flow prediction,'' {\em IEEE Transactions on
  Intelligent Transportation Systems (TITS)}, 2022.
\newblock doi: {10.1109/TITS.2022.3157056}.

\bibitem{caldarola2021cluster}
D.~Caldarola, M.~Mancini, F.~Galasso, M.~Ciccone, E.~Rodol{\`a}, and B.~Caputo,
  ``Cluster-driven graph federated learning over multiple domains,'' in {\em
  Proceedings of the IEEE/CVF Conference on Computer Vision and Pattern
  Recognition (CVPR)}, pp.~2749--2758, 2021.

\bibitem{DBLP:conf/aistats/McMahanMRHA17}
B.~McMahan, E.~Moore, D.~Ramage, S.~Hampson, and B.~A. y~Arcas,
  ``Communication-efficient learning of deep networks from decentralized
  data,'' in {\em Proceedings of the International Conference on Artificial
  Intelligence and Statistics (AISTATS)}, pp.~1273--1282, 2017.

\bibitem{wu2020comprehensive}
Z.~Wu, S.~Pan, F.~Chen, G.~Long, C.~Zhang, and S.~Y. Philip, ``A comprehensive
  survey on graph neural networks,'' {\em IEEE Transactions on Neural Networks
  and Learning Systemsv (TNNLS)}, vol.~32, no.~1, pp.~4--24, 2020.

\bibitem{DBLP:conf/iclr/KipfW17}
T.~N. Kipf and M.~Welling, ``Semi-supervised classification with graph
  convolutional networks,'' in {\em Proceedings of the International Conference
  on Learning Representations (ICLR)}, pp.~1--14, 2017.

\bibitem{DBLP:conf/iclr/VelickovicCCRLB18}
P.~Velickovic, G.~Cucurull, A.~Casanova, A.~Romero, P.~Li{\`{o}}, and
  Y.~Bengio, ``Graph attention networks,'' in {\em Proceedings of the
  International Conference on Learning Representations (ICLR)}, pp.~1--12,
  2018.

\bibitem{DBLP:conf/nips/HamiltonYL17}
W.~L. Hamilton, Z.~Ying, and J.~Leskovec, ``Inductive representation learning
  on large graphs,'' in {\em Advances in Neural Information Processing Systems
  (NIPS)}, pp.~1024--1034, 2017.

\bibitem{wang2020gcn}
X.~Wang, M.~Zhu, D.~Bo, P.~Cui, C.~Shi, and J.~Pei, ``Am-gcn: Adaptive
  multi-channel graph convolutional networks,'' in {\em Proceedings of the
  {ACM} {SIGKDD} International Conference on Knowledge Discovery \& Data Mining
  (SIGKDD)}, pp.~1243--1253, 2020.

\bibitem{zhong2023contrastive}
L.~Zhong, J.~Yang, Z.~Chen, and S.~Wang, ``Contrastive graph convolutional
  networks with generative adjacency matrix,'' {\em IEEE Transactions on Signal
  Processing (TSP)}, vol.~71, pp.~772--785, 2023.

\bibitem{sun2020multi}
K.~Sun, Z.~Lin, and Z.~Zhu, ``Multi-stage self-supervised learning for graph
  convolutional networks on graphs with few labeled nodes,'' in {\em
  Proceedings of the AAAI Conference on Artificial Intelligence},
  pp.~5892--5899, 2020.

\bibitem{tan2023federated}
Y.~Tan, Y.~Liu, G.~Long, J.~Jiang, Q.~Lu, and C.~Zhang, ``Federated learning on
  non-iid graphs via structural knowledge sharing,'' in {\em Proceedings of the
  AAAI Conference on Artificial Intelligence}, pp.~9953--9961, 2023.

\bibitem{wang2022federatedscope}
Z.~Wang, W.~Kuang, Y.~Xie, L.~Yao, Y.~Li, B.~Ding, and J.~Zhou,
  ``Federatedscope-gnn: Towards a unified, comprehensive and efficient package
  for federated graph learning,'' in {\em Proceedings of the ACM SIGKDD
  Conference on Knowledge Discovery and Data Mining (SIGKDD)}, pp.~4110--4120,
  2022.

\bibitem{he2021fedgraphnn}
C.~He, K.~Balasubramanian, E.~Ceyani, C.~Yang, H.~Xie, L.~Sun, L.~He, L.~Yang,
  P.~S. Yu, Y.~Rong, {\em et~al.}, ``Fedgraphnn: A federated learning system
  and benchmark for graph neural networks,'' {\em arXiv preprint
  arXiv:2104.07145}, 2021.

\bibitem{wu2021fedgnn}
C.~Wu, F.~Wu, Y.~Cao, Y.~Huang, and X.~Xie, ``Fedgnn: Federated graph neural
  network for privacy-preserving recommendation,'' {\em arXiv preprint
  arXiv:2102.04925}, 2021.

\bibitem{sen2008collective}
P.~Sen, G.~Namata, M.~Bilgic, L.~Getoor, B.~Galligher, and T.~Eliassi-Rad,
  ``Collective classification in network data,'' {\em AI Magazine}, vol.~29,
  no.~3, pp.~93--93, 2008.

\bibitem{mernyei2020wiki}
P.~Mernyei and C.~Cangea, ``Wiki-cs: A wikipedia-based benchmark for graph
  neural networks,'' {\em arXiv preprint arXiv:2007.02901}, 2020.

\bibitem{shchur2018pitfalls}
O.~Shchur, M.~Mumme, A.~Bojchevski, and S.~G{\"u}nnemann, ``Pitfalls of graph
  neural network evaluation,'' {\em arXiv preprint arXiv:1811.05868}, 2018.

\bibitem{blondel2008fast}
V.~D. Blondel, J.~Guillaume, R.~Lambiotte, and E.~Lefebvre, ``Fast unfolding of
  communities in large networks,'' {\em Journal of Statistical Mechanics:
  Theory and Experiment (JSTAT)}, vol.~2008, no.~10, p.~P10008, 2008.

\end{thebibliography}
	\newpage

\end{document}